\pdfoutput=1
\documentclass{article}

\usepackage[utf8]{inputenc} % allow utf-8 input
\usepackage[T1]{fontenc}    % use 8-bit T1 fonts

\usepackage[bitstream-charter]{mathdesign}
\usepackage{amsmath}
\usepackage[scaled=0.92]{PTSans}

\usepackage[
  paper  = letterpaper,
  left   = 1.15in,
  right  = 1.15in,
  top    = 1.0in,
  bottom = 1.0in,
  ]{geometry}

\usepackage[dvipsnames,table]{xcolor}
\definecolor{shadecolor}{gray}{0.9}

\usepackage[final,expansion=alltext]{microtype}
\usepackage[english]{babel}
\usepackage[parfill]{parskip}

\usepackage{graphicx}
\usepackage{placeins}
\usepackage[labelfont=bf,font=footnotesize,width=.9\textwidth]{caption}

\usepackage{booktabs,multirow,multicol}       % professional-quality tables

\usepackage[colorlinks,linktoc=all]{hyperref}
\usepackage[all]{hypcap}
\hypersetup{citecolor=MidnightBlue}
\hypersetup{linkcolor=MidnightBlue}
\hypersetup{urlcolor=MidnightBlue}

\usepackage[nameinlink]{cleveref}

\usepackage{amsthm}
\usepackage{mathtools}
\usepackage{thmtools}

\renewcommand{\epsilon}{\varepsilon}

\declaretheorem[style=plain,name=Theorem]{theorem}
\numberwithin{equation}{section}

\usepackage{csquotes}
\usepackage[%
minnames=1,maxnames=99,maxcitenames=2,minbibnames=10,maxbibnames=10,
style=alphabetic,
doi=false,
url=false,
giveninits=true,
hyperref,
natbib,
backend=bibtex,
sorting=nyt,
backref=true
]{biblatex}%
\renewbibmacro*{journal}{%
  \iffieldundef{journaltitle}
    {}
    {\printtext[journaltitle]{%
       \printfield[noformat]{journaltitle}%
       \setunit{\subtitlepunct}%
       \printfield[noformat]{journalsubtitle}}}}

\DeclareFieldFormat{sentencecase}{\MakeSentenceCase*{#1}}

\renewbibmacro*{title}{%
  \ifthenelse{\iffieldundef{title}\AND\iffieldundef{subtitle}}
    {}
    {\ifthenelse{\ifentrytype{article}\OR\ifentrytype{inbook}%
      \OR\ifentrytype{incollection}\OR\ifentrytype{inproceedings}%
      \OR\ifentrytype{inreference}\OR\ifentrytype{misc}}
      {\printtext[title]{%
        \printfield[sentencecase]{title}%
        \setunit{\subtitlepunct}%
        \printfield[sentencecase]{subtitle}}}%
      {\printtext[title]{%
        \printfield[titlecase]{title}%
        \setunit{\subtitlepunct}%
        \printfield[titlecase]{subtitle}}}%
     \newunit}%
  \printfield{titleaddon}}

\AtEveryBibitem{%
\ifentrytype{article}{
    \clearfield{urldate}%
    \clearfield{eprint}
    \clearfield{eid}
}{}
\ifentrytype{book}{
    \clearfield{url}%
    \clearfield{urldate}%
    \clearfield{eprint}
}{}
\ifentrytype{collection}{
    \clearfield{url}%
    \clearfield{urldate}%
    \clearfield{eprint}
}{}
\ifentrytype{incollection}{
    \clearfield{url}%
    \clearfield{urldate}%
    \clearfield{eprint}
}{}
}

\AtEveryBibitem{
    \ifentrytype{article}{}{\clearfield{pages}}%
    \clearfield{review}%
    \clearfield{series}%%
    \ifentrytype{article}{}{\clearfield{volume}}%
    \clearfield{month}
    \clearfield{isbn}
    \clearfield{issn}
    \clearlist{location}
    \clearfield{series}
    \clearlist{publisher}
    \clearname{editor}
}{}

\crefformat{equation}{(#2#1#3)}
\crefformat{figure}{Figure~#2#1#3}
\crefname{example}{Example}{Examples}
\crefname{lemma}{Lemma}{Lemmas}
\crefname{cor}{Corollary}{Corollaries}
\crefname{theorem}{Theorem}{Theorems}
\crefname{assumption}{Assumption}{Assumptions}

\usepackage{enumitem} % tight enumerates
\usepackage[separate-uncertainty=true,multi-part-units=single]{siunitx} % better table control

\usepackage[affil-it]{authblk}

\title{Optimization Dynamics Imprint Semantic Specificity in Contrastive Embedding Norms}
\date{}
\author{Ziwei Su}
\author{Junyu Ren}
\author{Victor Veitch}
\affil{University of Chicago\\
Correspondence: \href{mailto:suziwei@uchicago.edu}{\texttt{suziwei@uchicago.edu}}}

\begin{document}
\maketitle

\begin{abstract}
Contrastive embedding models trained with scale-invariant losses are typically paired with distance metrics like cosine similarity, effectively ignoring embedding magnitudes. However, surprisingly, empirical studies reveal that despite this, these ``discarded'' norms seem to correlate with semantic properties such as concept specificity, token frequency, and human uncertainty. In this work, we provide a formal theoretical framework explaining this phenomenon. By analyzing the optimization dynamics, we derive an analytic formula demonstrating that embedding length naturally encodes this information as a byproduct of the training process. We also show how this gives rise to signals that can serve as ``free'' calibration tools in specific models and retrieval tasks, providing a grounded explanation for a previously heuristic observation.
\end{abstract}

\section{Introduction}

Contrastive embedding models now span visual self-supervision (SimCLR, MoCo~\cite{chen2020simclr,he2020moco}), image-text pretraining (CLIP~\cite{radford2021clip}), and text retrieval (Sentence-BERT, E5~\cite{reimers2019sbert,wang2022e5}). Their objectives match positives against negatives with dot products or cosine similarities after mapping each embedding $z$ to $z/\|z\|$, making $\|z\|$ irrelevant to every training logit and to standard cosine inference. This design stabilizes logit scale, makes temperature meaningful, encourages hyperspherical geometry, and removes magnitude from the stated objective.

Yet embedding norm is repeatedly reported as a signal. Prior work links norm to confidence in self-supervised learning~\cite{draganov2025norms}, sample quality in face recognition~\cite{meng2021magface,kim2022adaface}, information gain in word embeddings~\cite{oyama2023norm}, and uncertainty in probabilistic contrastive learning~\cite{kirchhof2022nonisotropic,kirchhof2023probabilistic}. Retrieval results sharpen the puzzle: cosine-only scoring underestimates similarity for high-frequency words, and explicit norm discounting partly repairs the error~\cite{zhou2022problems,wannasuphoprasit2023solving}. Existing evidence therefore creates a tension: the loss is designed so $\|z\|$ cannot affect the optimized similarity, but trained models still encode useful information in $\|z\|$. What remains missing is a mechanism that predicts when norm grows, when it shrinks, and which direction a norm-aware correction should take.

The main contribution of this paper is to provide this missing mechanism. We show that the norm of a fixed input is governed by gradient-update heat and radial drift: the heat term captures inverse specificity and semantic ambiguity, while drift captures how the input fits between neighboring concepts. The resulting equilibrium equation predicts how these quantities set embedding length. Norm thus becomes a free inference-time signal: when pure cosine similarity is suboptimal, inverse-norm discounting can improve retrieval without retraining or extra supervision; when cosine similarity is nearly optimal, norm still provides information about answer quality and uncertainty that traditional benchmarks usually omit.

% \begin{itemize}
%     \item define $\tilde{\mathcal{V}}$ and $\tilde{\mathcal{D}}$ as that formalize specificity;
%     \item derive the stationary norm equation and identify its geometric interpretations;
%     \item verify these claims experimentally;
%     \item discuss cases where it helps classification and retrieval.
% \end{itemize}

\section{Related Work}

Work on hyperspherical representations explains why normalized contrastive objectives privilege directions. The standard reference point is the alignment--uniformity view of contrastive learning~\cite{wang2020alignment}, under which normalized objectives encourage models to shape angular structure on the unit sphere. Hyperspherical metric-learning papers such as NormFace~\cite{wang2017normface}, Heated-Up Softmax Embedding~\cite{zhang2018heated}, and Deep Metric Learning with Spherical Embedding~\cite{zhang2020spherical} also note that normalization induces inverse-norm gradients and nontrivial radial dynamics, which helps explain why the design works. Together, these results suggest that norm cannot simply be ignored, though predicting and interpreting it remains difficult.

\paragraph{Norm as an information carrier.} Several papers show that norm carries information even when the loss removes it from similarity scoring. Draganov et al.~\cite{draganov2025norms} prove that scale-invariant losses produce tangential gradients that make weight norms grow monotonically and correlate with confidence-like quantities. MagFace~\cite{meng2021magface} and AdaFace~\cite{kim2022adaface} use feature norm as a sample-quality proxy; Oyama et al.~\cite{oyama2023norm} connect word-embedding norm to information gain; Kirchhof et al.~\cite{kirchhof2022nonisotropic,kirchhof2023probabilistic} give probabilistic interpretations through von Mises--Fisher concentration and Monte Carlo InfoNCE. These works motivate radial information but leave open the mechanism we study: how specificity, uncertainty, and transmitted gradient drift jointly determine the equilibrium norm of an individual embedding.

\paragraph{Norm-aware similarity and geometry of models like CLIP.} Retrieval work shows where radial information changes rankings. Cosine-only matching understates similarity for high-frequency words, and norm discounting partly repairs the error~\cite{zhou2022problems,wannasuphoprasit2023solving}; cross-modal retrieval normalizations improve performance by suppressing hub-like gallery items~\cite{bogolin2022querybank}. In multimodal models, prior work documents modality gap, anisotropy, and nontrivial global geometry in CLIP despite normalized contrastive training~\cite{liang2022mind,tyshchuk2023isotropy,levi2025double}. We connect these phenomena to a stationary norm law: large norms can reflect uncertainty or hub structure rather than mere frequency, and the resulting correction should discount large norms rather than amplify them.

\section{Setup and Notation}

The theory tracks a contrastive embedding model which maps an input $x$ (a sentence in a text encoder, an image or caption in a vision-language model) to an embedding vector $z = f(x;\theta),$ which can be used to compare similarity between inputs.

\paragraph{Inputs and concepts.} Throughout, $x$ denotes such an input, and $\theta_t$ denotes the parameter of the model after $t$ optimizer steps. We study how the embedding $z_t(x):=f(x;\theta_t)$ changes as the parameters $\theta_t$ are updated during training. 

\paragraph{Architecture.} Typically, both the internal transformer layers and the final hidden state before the projection head undergo normalization (such as LayerNorm or RMSNorm), which re-centers and rescales representations to largely eliminate the effect of parameter magnitude within those stages. Consequently, the parameters \emph{after} this final normalization, namely the projection-head weights, exert the most direct control over the $\ell_2$ norm of the output embedding, while the rest of the network mainly governs direction rather than scale. For models without a projection head, the final layer weights play this role.

\paragraph{Training objective.} The canonical example is the symmetric InfoNCE objective used in contrastive multimodal models. Given a batch of $N$ matched pairs $(x_i, y_i)$ from two modalities, let $z_i^{(x)}$ and $z_j^{(y)}$ denote the unnormalized embeddings from the respective encoders, and $u_i = {z_i^{(x)}}/{\lVert z_i^{(x)} \rVert}, \quad v_j = {z_j^{(y)}}/{\lVert z_j^{(y)} \rVert}$. The similarity matrix is computed as $S_{ij} = u_i^\top v_j / \tau$, where $\tau > 0$ is a learned temperature parameter. The contrastive loss for the $x$-to-$y$ and $y$ to $x$ directions are:
\begin{equation}
\mathcal{L}_{x \to y} = -\frac{1}{N} \sum_{i=1}^{N} \log \frac{\exp(S_{ii})}{\sum_{j=1}^{N} \exp(S_{ij})}, \space \mathcal{L}_{y \to x} = -\frac{1}{N} \sum_{j=1}^{N} \log \frac{\exp(S_{jj})}{\sum_{i=1}^{N} \exp(S_{ij})}.
\end{equation}
The total loss is $\mathcal{L} = (\mathcal{L}_{x \to y} + \mathcal{L}_{y \to x})/2$. This loss depends on embeddings only through direction, not magnitude. Our theory applies to such scale-invariant loss.

\paragraph{Assumption on the loss.} InfoNCE is just a representative instance; throughout the paper we require only the following structural property of $\mathcal{L}$ (the specific form plays no role in the theory):

\begin{quote}
\textbf{(A)} \emph{Scale invariance.} For every $i$ and every $\alpha > 0$,
\[
\mathcal{L}(z_1, \ldots, \alpha z_i, \ldots, z_M) = \mathcal{L}(z_1, \ldots, z_i, \ldots, z_M).
\]
\end{quote}

Assumption~(A) holds whenever $\mathcal{L}$ computes similarities exclusively via $\ell_2$-normalized embeddings, as in InfoNCE and its variants. 

\begin{quote}
\textbf{(B)} \emph{Per-sample decomposition.} Given a minibatch $B_t = \{(x_k, y_k)\}_{k=1}^N$ of i.i.d.\ draws from $P_{\mathrm{data}}$, the loss decomposes as
\[
\mathcal{L} = \frac{1}{N} \sum_{k=1}^{N} \mathcal{L}_k,
\]
where each $\mathcal{L}_k$ depends on the pair $(x_k, y_k)$ and the other batch members only through the shared similarity scores.
\end{quote}

Assumption~(B) lets us write the parameter gradient as a sum of per-sample contributions, $\nabla_\theta \mathcal{L} = \frac{1}{N}\sum_k \nabla_\theta \mathcal{L}_k$, which is the starting point of the gradient-transmission analysis. (B) does \emph{not} require $\mathcal{L}_k$ to depend only on the $k$-th pair in isolation: as in InfoNCE, each $\mathcal{L}_k$ may depend on the full batch, provided the loss still averages such per-anchor terms.

\section{Weight Decay and Stationary Magnitude}\label{sec:theory}

Before the main theorem, consider a simple illustrative example. In most models, the final encoder layer projects the backbone feature vector $h_i \in \mathbb{R}^d$ for sample $i$ to the output embedding $z_i \in \mathbb{R}^k$. This layer is usually linear with weight matrix $W \in \mathbb{R}^{k \times d}$, so $z_i = W h_i$.
The contrastive loss depends only on normalized vectors, so it is independent of the scaling of $W$: $\mathcal{L}(\alpha W) = \mathcal{L}(W)$ for any scalar $\alpha > 0$.

Differentiating $\mathcal{L}(\alpha W) = \mathcal{L}(W)$ with respect to $\alpha$ at $\alpha=1$ gives\footnote{$\|\cdot\|_F$ denotes the Frobenius norm.}
\begin{equation}
\langle W, \nabla_W \mathcal{L} \rangle_F = \text{Tr}(W^\top \nabla_W \mathcal{L}) = 0.
\end{equation}
Under a gradient descent update $W' = W - \eta \nabla_W \mathcal{L}$, the squared Frobenius norm of the weights evolves as:
\begin{align}
\|W'\|_F^2 &= \|W - \eta \nabla_W \mathcal{L}\|_F^2 = \|W\|_F^2 - 2\eta \langle W, \nabla_W \mathcal{L} \rangle_F + \eta^2 \|\nabla_W \mathcal{L}\|_F^2.\\
&=\|W\|_F^2 + \eta^2 \|\nabla_W \mathcal{L}\|_F^2,
\end{align}

Thus the magnitude grows whenever a nonzero gradient step lies orthogonal to $W$. Qualitatively similar radial-growth effects due to tangential gradients were also noted in~\cite{wang2017normface,zhang2018heated,draganov2025norms}. 
However, the main purpose of this paper is to derive a formula for the length of \emph{individual} inputs.

Weight decay is usually applied to prevent norms from growing too large. Modern implementations such as AdamW do weight decay as a separate update from the gradient step. Let $U_t$ denote the task step before decay (for SGD, $U_t=-\eta G_t$), and let $\Lambda$ be the parameter-wise decay operator (diagonal, with entry $\lambda_j$ for each parameter, with entries zero for parameters not decayed). A decoupled-weight-decay update can be split into an intermediate task state and a final decay step,
\begin{equation}
\label{eq:decoupled_wd}
\theta_t^+=\theta_t+U_t,
\qquad
\theta_{t+1}=\theta_t^+ + \Delta\theta_{\mathrm{wd}}.
\end{equation}
For the usual AdamW ordering, $\Delta\theta_{\mathrm{wd}}=-\eta\Lambda\theta_t$; for an explicit task-then-decay ordering, $\Delta\theta_{\mathrm{wd}}=-\eta\Lambda\theta_t^+$. The analysis below only uses the induced post-task decay step, so either convention is represented by the measured effective damping.

After training, the task update does not necessarily vanish; instead, the process reaches a stationary distribution. For a fixed input $x$, define
\[
\begin{aligned}
z_t^+(x)&:=f(x;\theta_t^+),\\
\Delta z_{\mathrm{task}}(x) := f(x;\,\theta_t+U_t) - f(x;\,\theta_t),
\qquad
\Delta z_{\mathrm{wd}}(x) := f(x;\,\theta_{t+1}) - f(x;\,\theta_t^+),
\end{aligned}
\]
so that $\Delta z_t = \Delta z_{\mathrm{task}} + \Delta z_{\mathrm{wd}}$ exactly. The effective radial damping is defined by the squared-norm contraction caused by the post-task decay step:
\begin{equation}
\label{eq:lambda_rad_stationary}
\mathbb{E}_{\pi,B_t}\!\left[
\|z_{t+1}(x)\|^2-\|z_t^+(x)\|^2
\,\middle|\,\|z_t(x)\|=R
\right]
=
-2\eta\bar\lambda_{\mathrm{rad}}(x;R)R^2.
\end{equation}
Equivalently, this definition includes
\[
2\bigl(z_t(x)+\Delta z_{\mathrm{task}}(x)\bigr)^\top\Delta z_{\mathrm{wd}}(x)
+\|\Delta z_{\mathrm{wd}}(x)\|^2
\]
inside the effective damping. Here $\pi$ denotes the stationary distribution of the training state, including parameters and any optimizer state; minibatch randomness is averaged separately through $B_t$. When this effective damping varies slowly over the stationary range of $R$, we write it as $\bar\lambda_{\mathrm{rad}}(x)$. The condition needed below is a positive net time-averaged contraction, not pointwise positivity of the decay contribution at every step.

\subsection{The Main Equation of Equilibrium Length}
When training converges to a stationary distribution, the squared norm has zero drift after averaging over that stationary distribution. The closed-form norm below is the corresponding zero-drift radius under a narrow stationary norm distribution and slowly varying coefficients.
\begin{theorem}[Stationary Embedding Norm]
\label{thm:master_norm}
Let $x$ be a fixed input whose embedding $z_t(x)=f(x;\theta_t)$ is trained under a loss function satisfying assumptions (A), (B) with learning rate $\eta>0$ and decoupled weight decay. Assume its stationary effective radial damping $\bar\lambda_{\mathrm{rad}}(x;R)$ is positive for the radii under consideration, in particular near the zero-drift radius defined below. At each step $t$, given the parameters $\theta_t$, the minibatch $B_t$ is random; hence $\Delta z_{\mathrm{task}}(x)$ is a random vector. Define the local, state-conditioned task moments
\[
\begin{aligned}
\Delta\tilde z_{\mathrm{task}}(x)&:=\|z_t(x)\|\Delta z_{\mathrm{task}}(x),\\
\mu_t(x) &:= \mathbb{E}_{B_t}[\Delta\tilde z_{\mathrm{task}}(x)\mid \theta_t],\\
\tilde{\mathcal{V}}_t(x)^2 &:= \mathrm{Tr}\!\left(\mathrm{Cov}_{B_t}(\Delta\tilde z_{\mathrm{task}}(x)\mid \theta_t)\right),
\end{aligned}
\]
and let $\hat{z}_t(x):=z_t(x)/\|z_t(x)\|$ be its unit direction. Write $R_t:=\|z_t(x)\|$. Define the radius-conditioned stationary averages
\begin{align}
\tilde{\mathcal{D}}_R(x)
&:= \mathbb{E}_{\pi}\!\left[\hat z_t(x)^\top\mu_t(x)\,\middle|\,\|z_t(x)\|=R\right],\\
\tilde{\mathcal{V}}_R(x)^2
&:= \mathbb{E}_{\pi}\!\left[\tilde{\mathcal{V}}_t(x)^2\,\middle|\,\|z_t(x)\|=R\right],\\
\mathcal{M}_R(x)^2
&:= \mathbb{E}_{\pi}\!\left[\|\mu_t(x)\|^2\,\middle|\,\|z_t(x)\|=R\right],\\
\Sigma_R(x)^2&:=\mathcal{M}_R(x)^2+\tilde{\mathcal{V}}_R(x)^2.
\end{align}
Given $\|z_t(x)\|=R$, the task step alone changes the squared norm by
\begin{equation}
\mathbb{E}_{\pi,B_t}\!\left[\|z_t^+(x)\|^2 - R^2 \,\middle|\, \|z_t(x)\|=R\right]
= 2\tilde{\mathcal{D}}_R(x) + \frac{\Sigma_R(x)^2}{R^2}.
\end{equation}
Combining this task contribution with the effective post-task decay contraction in Eq.~\eqref{eq:lambda_rad_stationary} gives the radius-conditioned drift
\begin{equation}
\mathbb{E}_{\pi,B_t}\!\left[\|z_{t+1}(x)\|^2 - R^2 \,\middle|\, \|z_t(x)\|=R\right]
= 2\tilde{\mathcal{D}}_R(x) + \frac{\Sigma_R(x)^2}{R^2} - 2\eta\bar\lambda_{\mathrm{rad}}(x;R)R^2.
\end{equation}
Stationarity implies that the outer average of this radius-conditioned drift vanishes,
\begin{equation}
0
=
\mathbb{E}_{\pi}\!\left[
2\tilde{\mathcal{D}}_{R_t}(x)
+\frac{\Sigma_{R_t}(x)^2}{R_t^2}
-2\eta\bar\lambda_{\mathrm{rad}}(x;R_t)R_t^2
\right].
\end{equation}
To obtain a single representative norm, assume the stationary norm distribution is concentrated near a zero-drift radius $R_{\mathrm{eq}}$ and that the radius-conditioned quantities vary slowly there; equivalently, define $R_{\mathrm{eq}}$ as a radius where the displayed radius-conditioned drift is zero. At $R=R_{\mathrm{eq}}$, define
\begin{equation}
\mathcal{D}_\star(x)
:=
\frac{\tilde{\mathcal{D}}(x)}{\eta\bar\lambda_{\mathrm{rad}}(x)},
\qquad
\mathcal{V}_\star(x)^2
:=
\frac{\Sigma(x)^2}{\eta\bar\lambda_{\mathrm{rad}}(x)}.
\end{equation}
The zero-drift condition gives the bi-quadratic balance
\begin{equation}
\label{eq:master_biquadratic}
2 R_{\mathrm{eq}}^4 - 2\mathcal{D}_\star(x)\, R_{\mathrm{eq}}^2 - \mathcal{V}_\star(x)^2 = 0,
\end{equation}
whose nonnegative zero-drift solution is
\begin{equation}
\label{eq:master_norm}
R_{\mathrm{eq}}^2 = \frac{\mathcal{D}_\star(x) + \sqrt{\mathcal{D}_\star(x)^2 + 2\mathcal{V}_\star(x)^2}}{2}.
\end{equation}
When $\mathcal{V}_\star(x)>0$, this solution is strictly positive and unique.
\end{theorem}

\paragraph{What the variables measure.}
The theorem separates two scale-normalized task-update forces after stationary time averaging. $\tilde{\mathcal{V}}(x)^2$ is stochastic task heat: it is large when random minibatches send mutually inconsistent update votes. $\tilde{\mathcal{D}}(x)$ is structural radial drift.

\paragraph{The bi-quadratic Eq.~\eqref{eq:master_biquadratic} yields three interpretable limits.} \textbf{Variance-dominated} concepts ($|\mathcal{D}_\star(x)| \ll \mathcal{V}_\star(x)$) satisfy $R_{\mathrm{eq}} \approx (\mathcal{V}_\star(x)^2/2)^{1/4}$; if the radius-conditioned mean-update energy $\mathcal{M}(x)^2 \ll \tilde{\mathcal{V}}(x)^2$, then $R_{\mathrm{eq}} \approx (\tilde{\mathcal{V}}(x)^2/(2\eta\bar\lambda_{\mathrm{rad}}(x)))^{1/4}$. Ambiguous concepts often live here: no stable neighborhood pulls them outward, but stochastic update heat still resists decay. \textbf{Outward-drift} concepts ($\mathcal{D}_\star(x)>0$, $\mathcal{D}_\star(x)^2\gg\mathcal{V}_\star(x)^2$) have $R_{\mathrm{eq}}^2\approx\mathcal{D}_\star(x)$ and $R_{\mathrm{eq}} \approx (\tilde{\mathcal{D}}(x)/(\eta\bar\lambda_{\mathrm{rad}}(x)))^{1/2}$. Broad semantic hubs live here when many neighbors transmit updates that agree with the radial direction of $x$. \textbf{Inward-drift} concepts ($\mathcal{D}_\star(x)<0$) suffer extra radial damping; in the strong inward limit $|\mathcal{D}_\star(x)|^2\gg 2\mathcal{V}_\star(x)^2$, rationalizing the positive root gives $R_{\mathrm{eq}}^2 \approx \mathcal{V}_\star(x)^2/(2|\mathcal{D}_\star(x)|) = (\mathcal{M}(x)^2+\tilde{\mathcal{V}}(x)^2)/(2|\tilde{\mathcal{D}}(x)|)$. Concepts whose positive contexts span incompatible embedding directions can live here.

\subsection{NTK approximation and specificity}
The NTK helps explain why $\tilde{\mathcal{V}}$ and $\tilde{\mathcal{D}}$ measure these forces. The following local expressions estimate the state-conditioned moments $\mu_t$ and $\tilde{\mathcal{V}}_t^2$; their radius-conditioned stationary averages give the theorem quantities. Let $s_k:=\|z_t(X_k)\|\nabla_{z_t(X_k)}\mathcal{L}$ be the scale-normalized source gradient for the $k$-th batch element\footnote{Here $\mathcal{L}$ is the full minibatch loss. The notation $\nabla_{z_t(X_k)}\mathcal{L}$ means the gradient of the full minibatch loss with respect to the embedding in slot $k$, with the other batch embeddings held fixed for this partial derivative. It is not the gradient of an isolated single-example loss: InfoNCE-style denominators couple all batch items.}, with $\bar{s}(X)=\mathbb{E}[s_k\mid X_k=X]$ and $\Sigma_s(X)=\mathrm{Cov}(s_k\mid X_k=X)$. Since these source gradients are tangent and inverse-length scaled, neighboring embeddings have similar norms $R\approx\|z_t(x)\|\approx\|z_t(X)\|$, and the minibatch averaging is already included in the full-loss gradient defining $s_k$, the chain rule $\Delta z = J\Delta\theta$ gives $\Delta \tilde z_{\mathrm{task}} \approx -\eta\sum_{k\in B_t}\mathcal{K}(x,X_k)s_k$. For exchangeable batch slots, neglecting cross-slot covariance terms induced by contrastive denominators gives the moment approximation (see Appendix~\ref{app:variance_decomp} for derivation):
\begin{align}
\mu_t(x)
&\approx
-\eta B\,\mathbb{E}_{X}\!\left[\mathcal{K}(x,X)\bar{s}(X)\right], \\
\tilde{\mathcal{V}}_t(x)^2
&\approx
\eta^2 B\,\mathrm{Tr}\!\left(
\mathbb{E}_{X}\!\left[\mathcal K(x,X)\Sigma_s(X)\mathcal K(x,X)^\top\right]
+\mathrm{Cov}_{X}\!\left(\mathcal K(x,X)\bar{s}(X)\right)
\right).
\end{align}
These NTK expressions are an interpretive decomposition of the task moments, not the computational estimator used in the experiments. The empirical protocol estimates $\tilde{\mathcal{D}}$ and $\tilde{\mathcal{V}}^2$ directly from virtual task updates, while the displayed formulas explain how those moments can arise from NTK neighborhood transmission under the weak-coupling approximation above.
When the conditional mean is small, the variance proxy reduces to
\begin{equation}
\label{eq:energy_scaling}
\tilde{\mathcal{V}}_t(x)^2
\approx
\eta^2 B\,\mathbb{E}_{X}\!\left[
s(X)^\top\mathcal K(x,X)^\top\mathcal K(x,X)s(X)
\right].
\end{equation}
Thus the local $\tilde{\mathcal{V}}_t$ and its radius-conditioned stationary average $\tilde{\mathcal{V}}$ are large when the ``NTK neighborhood'' of $x$ (inputs treated as similar to $x$ by the model at parameter $\theta_t$) contains mutually inconsistent information according to training labels. Generic concepts, polysemous inputs, and perceptually ambiguous images receive diverse tangent kicks; specific concepts receive small, repeatable kicks because their positives concentrate around one direction. In regimes where $\tilde{\mathcal{V}}$ increases with $R$, norm becomes a proxy for inverse specificity.

\begin{figure}[t]
  \centering
  \includegraphics[width=0.55\linewidth]{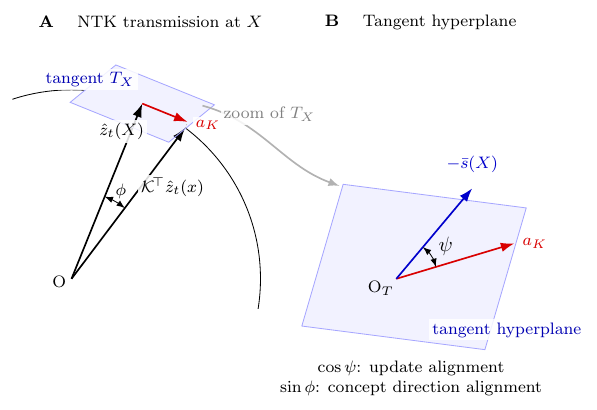}
  \caption{Sphere geometry of the radial drift term. At a training sample $X$, the transported radial direction $\mathcal K(x,X)^\top\hat z(x)$ has tangent projection $a_K$ in $T_X\mathbb S^{d-1}$. The source update $-\bar{s}(X)$ also lies in this tangent space. Its angle $\psi_K$ with $a_K$ sets the sign of the radial vote: $\cos\psi_K>0$ pushes $x$ outward, while $\cos\psi_K<0$ pushes it inward.}
  \label{fig:radial_velocity_angles}
\end{figure}

Radial drift depends on the sign of those transmitted votes. Since the source gradient is tangent to the unit sphere at $X$, only the tangent projection
\[
a_K(x,X)
=
P_{\hat z(X)}^\perp\mathcal K(x,X)^\top\hat z(x)
\]
can change the length of $z(x)$. If $\varphi_K(x,X)$ is the angle between $\mathcal K(x,X)^\top\hat z(x)$ and $\hat z(X)$, and $\psi_K(x,X)$ is the angle between $a_K(x,X)$ and $-\bar{s}(X)$, then
\begin{equation}
\label{eq:drift_interp_projection}
\tilde{\mathcal{D}}_t(x)
\approx
\eta B\,\mathbb{E}_{X}\!\left[
\|\mathcal K(x,X)^\top\hat z(x)\|
\sin\varphi_K(x,X)
\|\bar{s}(X)\|
\cos\psi_K(x,X)
\right].
\end{equation}
Positive drift requires both strong transmission and positive alignment. A frequent but NTK-isolated input has small $\|a_K(x,X)\|$ and stays variance-dominated. A genuine hub has large $\|a_K(x,X)\|$ for many neighbors and $\cos\psi_K>0$ on average, so neighborhood updates lengthen its vector. A boundary concept can have equally strong transmission but $\cos\psi_K<0$, so the same NTK coupling compresses the norm. Figure~\ref{fig:radial_velocity_angles} illustrates the geometry. The source gradient $\bar{s}(X)$ is tangent to the sphere because the loss depends only on the normalized embedding $\hat{z}=z/\|z\|$; scale invariance gives $\nabla_z\mathcal{L}\cdot z=0$.

\subsection{Implications for Norm-Aware Similarity Scoring}
\label{sec:scoring}

The task-heating term $\mathcal{V}_\star(x)$ encodes neighborhood uncertainty: low $\mathcal{V}_\star$ marks a specific, well-defined concept, so a match between two low-$\mathcal{V}_\star$ inputs carries more information at the same cosine angle than a match between two high-$\mathcal{V}_\star$ inputs.

At inference, it is possible to compute $\mathcal{V}_\star$ at some cost but $R$ is already present. Theorem \ref{thm:master_norm} makes $R$ a possible proxy for $\mathcal{V}_\star$ when variance dominates. This then motivates a norm-discounted similarity score
\begin{equation}
\label{eq:norm_aware_score}
\mathrm{score}(q, d) = \cos\theta(q,d) \cdot \|q\|^{-\gamma} \cdot \|d\|^{-\gamma},
\end{equation}
where $\gamma > 0$ discounts high-norm, high-uncertainty embeddings. Pure cosine corresponds to $\gamma = 0$ and discards radial information; the dot product corresponds to $\gamma = -1$, amplifying the embeddings this argument says to downweight. 

\subsection{Optimizer Effects on Frequency Scaling (AdamW vs. SGD)}
We derive the main equation for SGD-style updates, but the core equilibrium principle does not depend on this optimizer choice. Empirically, $\tilde{\mathcal{V}}$ computed with virtual SGD steps remains informative for models trained with other optimizers. The math changes in predictable ways: for example, momentum rescales stochastic heating by $(1-\beta)/(1+\beta)$ without changing which concepts receive larger $\tilde{\mathcal{V}}$ (Appendix~\ref{app:optimizer}). AdamW combines this momentum rescaling with the decoupled weight decay analyzed in Section~\ref{sec:theory}.

\section{Experiments}
\label{sec:experiments}

We verify the main equation's predictions, then test when the induced norm signal improves similarity scoring.

\subsection{Verifying the main equation: equilibrium scaling with \texorpdfstring{$\lambda$}{lambda}}
\label{sec:exp_dv_equation}

\begin{figure}[t]
  \centering
  \begin{minipage}[t]{0.48\linewidth}
    \centering
    \includegraphics[width=\linewidth]{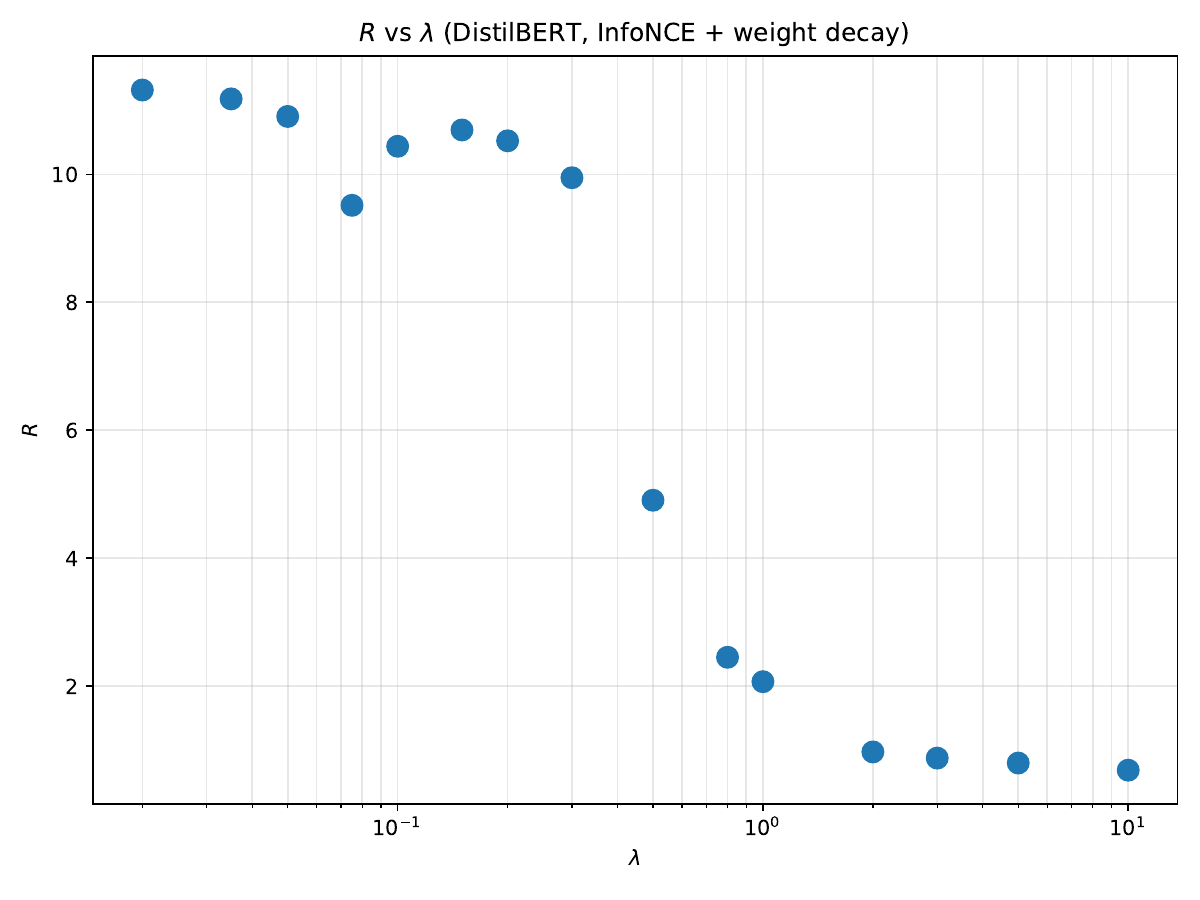}
  \end{minipage}
  \hfill
  \begin{minipage}[t]{0.48\linewidth}
    \centering
    \includegraphics[width=\linewidth]{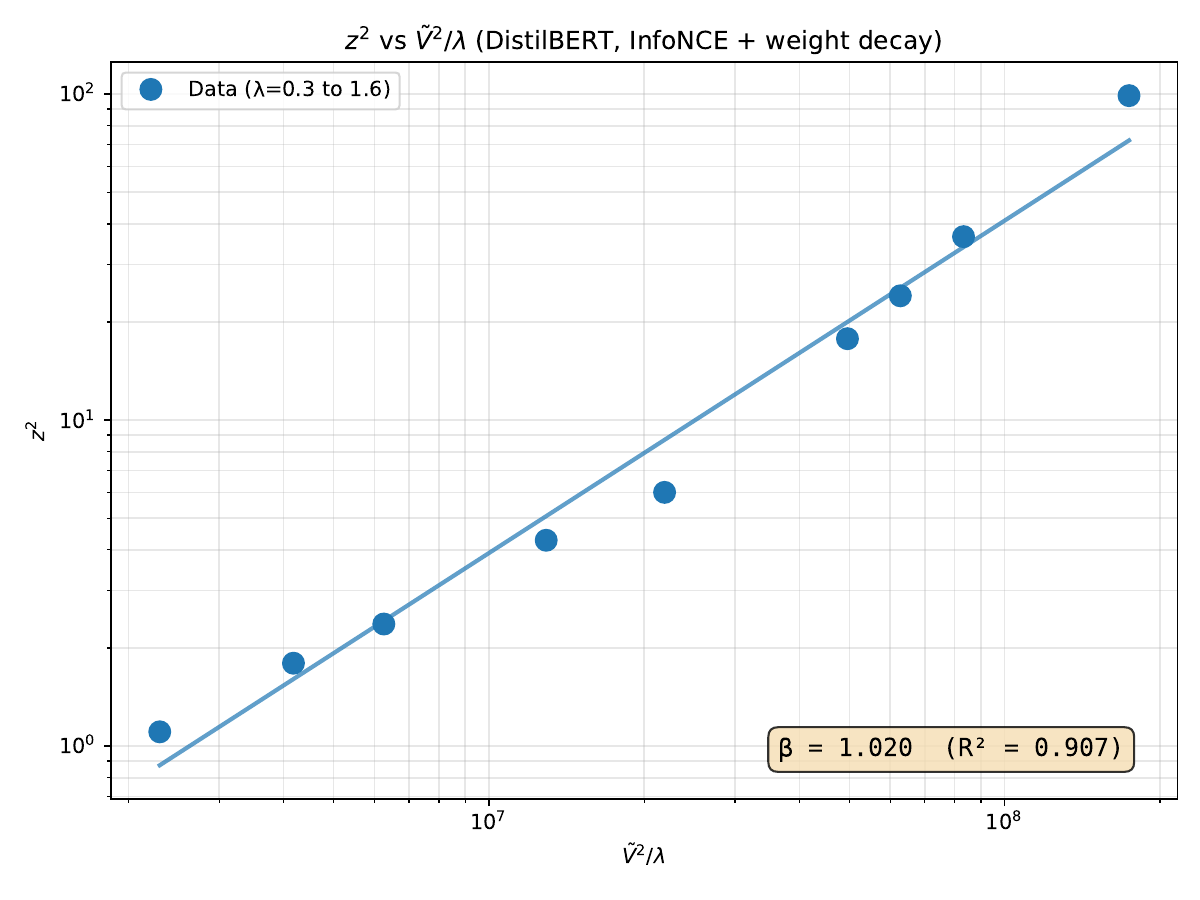}
  \end{minipage}
  \caption{Left: average $R$ across tokens against $\lambda$ on log scale. Right: log-log fitting of $R^4$ against $\tilde{\mathcal{V}}^2/\lambda$ supporting the linearized theoretical scaling.}
  \label{fig:norm_vs_lambda}
\end{figure}

\paragraph{$\lambda$ scaling.} The variance-dominated limit predicts $R_{\mathrm{eq}}^4 \propto \tilde{\mathcal{V}}^2/\lambda$, equivalently $R_{\mathrm{eq}}^2 \propto (\tilde{\mathcal{V}}^2/\lambda)^{1/2}$, coupling equilibrium norm to task heat and weight decay. To verify this relation, we fine-tune \texttt{distilbert-base-uncased}~\cite{sanh2019distilbert} on a synthetic dataset containing artificial concepts, each paired with 100 positive contexts consisting of diverse sentence variations. Using this set, we train the model (initialized with the original weights of DistilBERT) with SGDW over a range of decay values $\lambda$, training each run to norm convergence. $R$ decreases as $\lambda$ increases (Figure~\ref{fig:norm_vs_lambda}, left). The right panel shows that the linearized predicted scaling also holds: a log-log linear fit of $R^4$ against $\tilde{\mathcal{V}}^2/\lambda$ gives a slope close to $1$, verifying the coupled relationship introduced by the main equation.

\paragraph{Verifying the main equation.} To verify Theorem~\ref{thm:master_norm}, we fine-tune DistilBERT~\cite{sanh2019distilbert} on MultiNLI~\cite{williams2018mnli} using SGDW ($\eta=0.01$, $w=0.01$). We measure the norms of final mean-pooled embeddings for single-token inputs of the form \texttt{[CLS]} $w$ \texttt{[SEP]} at convergence, where $w$ ranges over selected vocabulary tokens, and estimate $\tilde{\mathcal{V}}$ and $\tilde{\mathcal{D}}$ via a Monte Carlo estimator. Because the pointwise radial decay varies across vocabulary positions and training steps, we estimate the effective stationary average. Details are in Appendix~\ref{app:dv_measurement}.

Figure~\ref{fig:predicted_vs_actual} presents the results. Panel~(a) tests the full equation: plotting the predicted norm against the observed norm yields a log-log gradient of $0.88$ with Pearson correlation $r=0.97$. The gradient $0.88 < 1$ and the y-intercept reflect a systematic under-prediction that stems from residual non-stationarity: although the model has reached convergence by any practical standard (training loss is flat, gradients are small), the embedding norms are still slowly decreasing over the measurement window. This biases the prediction slightly downward, but the high correlation shows that the balance equation captures the dominant mechanism, and the bias magnitude is quantitatively consistent with the measured residual drift.

Because $\tilde{\mathcal{D}}$ and $\bar\lambda_\mathrm{rad}$ are noisy to estimate precisely (for $\tilde{\mathcal{D}}$, estimators disagree on sign for 85\% of tokens; for $\bar\lambda_\mathrm{rad}$, the median within-token CV across 20 repeated measurements is 0.64), it is worth examining whether the noise-dominated limit, which only involves $\tilde{\mathcal{V}}$, holds in practice. When $\tilde{\mathcal{D}} \approx 0$, Eq.~\eqref{eq:master_norm} reduces to $\|z_\mathrm{eq}\| \approx (\tilde{\mathcal{V}}^2 / 2\eta\bar\lambda_\mathrm{rad})^{1/4}$, predicting a log-log slope of $0.25$ between $\tilde{\mathcal{V}}^2$ and $\|z_\mathrm{eq}\|$. Panel~(b) confirms this: the measured slope is $0.26$ ($r=0.95$). In contrast to $\tilde{\mathcal{D}}$ and $\bar\lambda_\mathrm{rad}$, the between-estimator correlation for $\Sigma^2$ (from which $\tilde{\mathcal{V}}^2$ is derived) is $r=0.91$, and $\tilde{\mathcal{V}}^2$ predicts the norm at $r=0.95$ across tokens. This robustness makes $\tilde{\mathcal{V}}$ a practical tool for analyzing embedding geometry across diverse models without retraining, as we will see in later subsections.

\begin{figure}[t]
  \centering
  \includegraphics[width=0.75\linewidth]{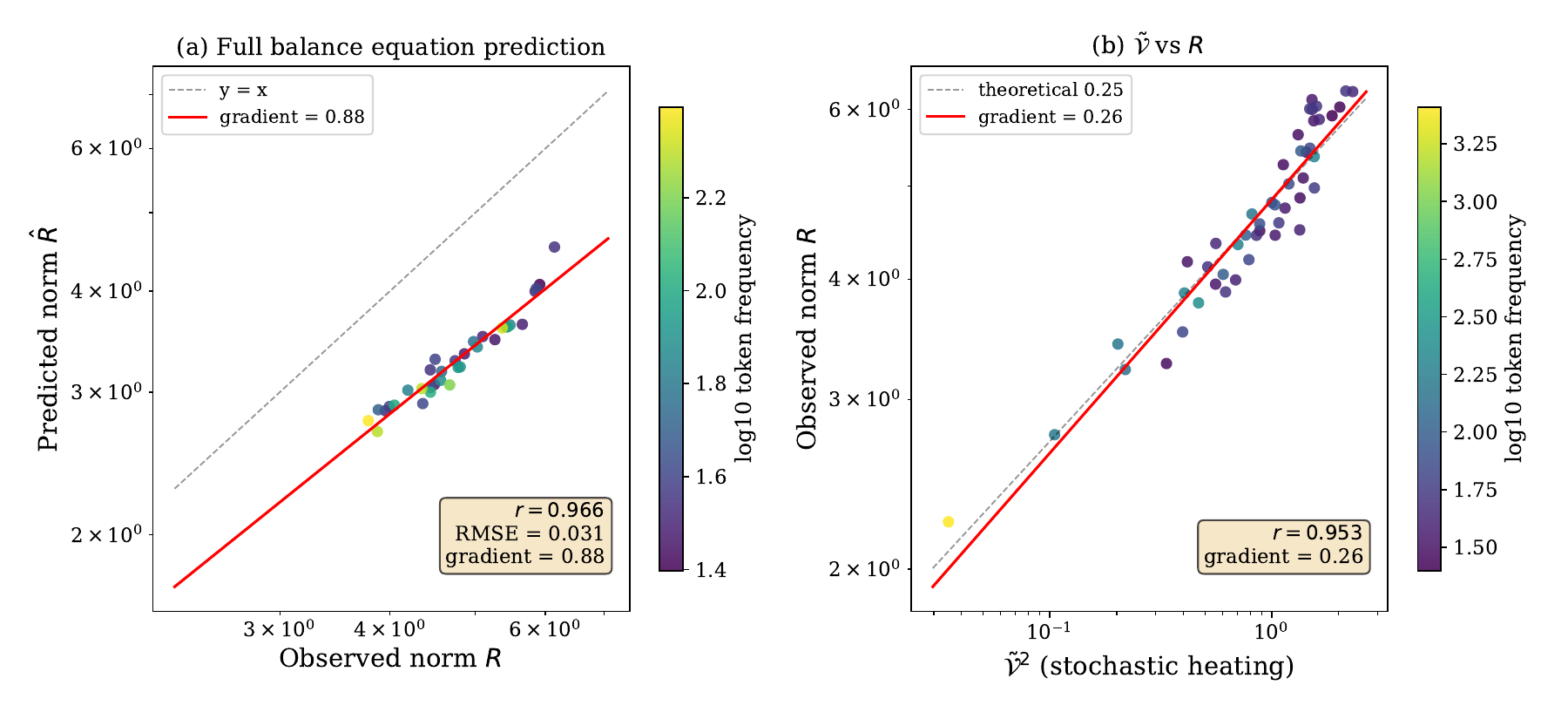}
  \caption{\textbf{(a)} Log-log scatter of predicted vs.\ observed embedding norm from the full balance equation (Eq.~\eqref{eq:master_norm}). Each point is a vocabulary position; color encodes token frequency. The fitted gradient is $0.88$ ($r=0.97$), showing that $\tilde{\mathcal{V}}$ and $\bar\lambda_\mathrm{rad}$ capture the dominant mechanism governing equilibrium norms. The systematic deviation from $y=x$ reflects residual non-stationarity in the measurement window. \textbf{(b)} $\tilde{\mathcal{V}}^2$ vs.\ $\|z_\mathrm{eq}\|$ on log-log axes. The fitted gradient $0.26$ matches the theoretical $+0.25$ prediction (dashed), supporting stochastic heating as the primary driver of embedding length.}
  \label{fig:predicted_vs_actual}
\end{figure}

In Appendix~\ref{app:more_models_verification}, we repeat this analysis for CLIP~\cite{radford2021clip} text embeddings and MiniLM-L12~\cite{wang2020minilm} \emph{without} fine-tuning and observe the same effect, though more weakly. Thus, although the quantities depend on optimizers and training data, the measured signal persists across model families.

\subsection{Frequency Independence}
\label{sec:freq_independence}

A natural objection is that norm might merely reflect training frequency rather than specificity. The main equation contains no independent count variable: the number of positive contexts $\rho$ enters only through $\mathcal D_{\star,\rho}(x)$ and $\mathcal V_{\star,\rho}(x)^2$, and its effect saturates rapidly. 

We verify this by training several architectures on synthetic datasets where $\rho$ varies from 5 to 2000 while holding semantic content fixed. Norms are essentially flat---the per-model coefficient of variation stays under $2\%$ across two orders of magnitude of $\rho$ (Table~\ref{tab:freq_independence}). 
\begin{table}[t]
\caption{Embedding norm is independent of concept frequency across architectures. Norms are reported at each $\rho$ (number of positive contexts) for a fixed concept; values are stable across two orders of magnitude of $\rho$ (max CV $<2\%$). Training details: mixed-$\rho$ SGD with weight decay, no frozen parameters.}
\label{tab:freq_independence}
\centering
\begin{tabular}{lccccc}
\toprule
Model & $\rho{=}5$ & $\rho{=}20$ & $\rho{=}100$ & $\rho{=}500$ & $\rho{=}2000$ \\
\midrule
  DistilBERT (base-uncased) & $7.346$ & $7.307$ & $7.294$ & $7.303$ & $7.344$ \\
  E5 (intfloat/e5-base) & $2.929$ & $2.934$ & $2.953$ & $2.829$ & $2.893$ \\
  SigLIP (base-patch16-384) & $4.188$ & $4.189$ & $4.191$ & $4.195$ & $4.173$ \\
  CLIP (ViT-B/32) & $23.765$ & $23.768$ & $23.867$ & $23.587$ & $22.946$ \\
\bottomrule
\end{tabular}
\end{table}

\subsection{Norm-aware scoring improves retrieval in specific regimes}
\label{sec:retrieval}

Section~\ref{sec:scoring} proposed the norm-discounted similarity score
\begin{equation*}
\mathrm{score}(q, d) = \cos\theta(q,d)\cdot\|q\|^{-\gamma}\cdot\|d\|^{-\gamma}.
\end{equation*}
For a fixed query, $\|q\|^{-\gamma}$ is constant and cannot affect the within-query ranking, so retrieval experiments use the equivalent document-weighted score $\cos(q,d)\cdot\|d\|^\alpha$ with $\alpha=-\gamma$. The symmetric score remains relevant for pairwise calibration across varying queries. In retrieval settings that prefer specific answers, $\alpha<0$ discounts high-norm, generic documents, whereas the dot product corresponds to $\alpha=1$ and boosts them.

\begin{figure}[t]
  \centering
  \includegraphics[width=\linewidth]{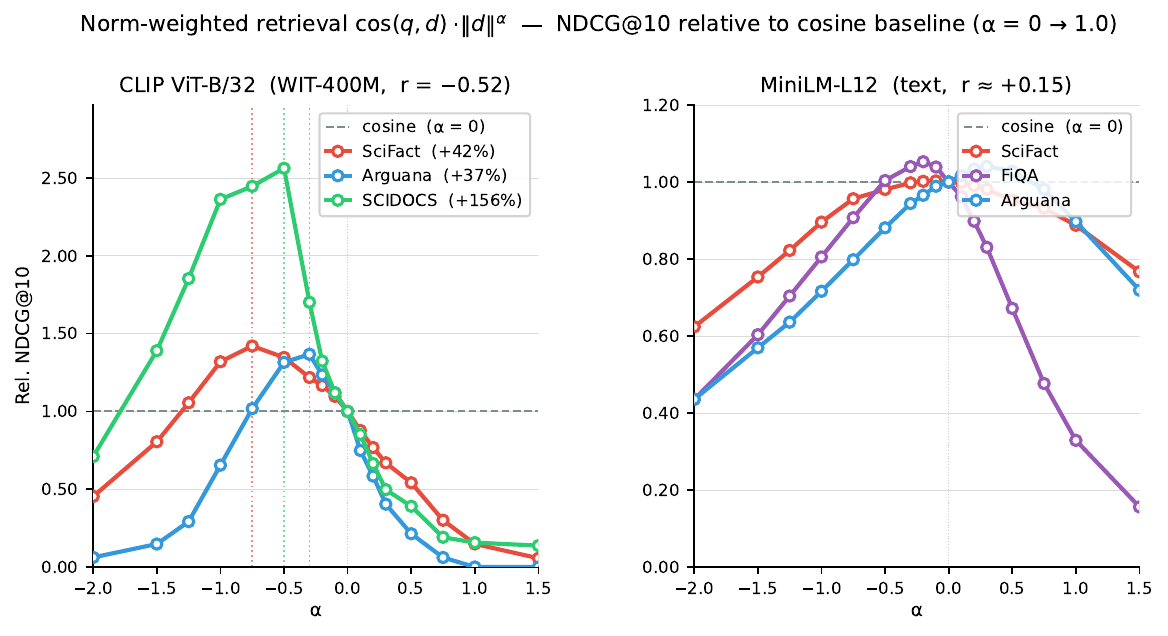}
  \caption{Norm-weighted BEIR retrieval: $\mathrm{score}(q,d)=\cos(q,d)\cdot\|d\|^\alpha$ ($\alpha=-\gamma$). NDCG@10 normalized to cosine baseline ($\alpha=0\to1.0$). \textbf{Left:} CLIP ViT-B/32 gains 37--156\% at negative $\alpha$ across three datasets. \textbf{Right:} MiniLM-L12 shows no benefit at any $\alpha$.}
  \label{fig:retrieval_curve}
\end{figure}

Figure~\ref{fig:retrieval_curve} shows a sweep on BEIR~\cite{thakur2021beir}, a heterogeneous benchmark of information retrieval datasets, with the document exponent $\alpha$. The $\alpha$ sweep is diagnostic: it shows whether norm can improve rankings and in which direction, rather than selecting a fixed deployment value. Although CLIP is not designed for text-to-text retrieval, norm discounting yields large gains, suggesting that norm carries useful information beyond what the training objective explicitly optimizes. Appendix~\ref{app:trec_car} gives a complementary depth-stratified TREC CAR comparison across additional models; the text-only models show smaller and model-dependent benefits there.

\section{Conclusion}

This paper provides a theoretical mechanism explaining why embedding norms in contrastive models correlate with semantic properties despite being discarded by the training loss. Our central result, Theorem~\ref{thm:master_norm}, presents a bi-quadratic equilibrium equation that balances radial weight decay against two update statistics: structural radial drift ($\tilde{\mathcal{D}}$) and stochastic heating ($\tilde{\mathcal{V}}$). In heat-controlled settings, $\tilde{\mathcal{V}} \propto R^2$, making the norm a free inference-time proxy for inverse semantic specificity and labeling uncertainty. Building on this insight, we demonstrate that a norm-aware scoring function that discounts large norms improves retrieval performance in the projection-head regimes where the theory predicts a usable norm signal, requiring no additional training or supervision.

\section{Limitations}

The main caveat is that the loss-function assumptions are not entirely consistent with mean pooling: scaling one token state before pooling can change the pooled direction, so the current equation is not quantitatively exact for mean-pooled embeddings. Mean pooling shows similar qualitative radial-heating and norm-specificity behavior, suggesting that a pooling-aware generalization would substantially broaden the theory. The same applies to optimizer choice: as discussed above, the derivation uses SGD-style updates, while momentum and AdamW change the effective drift and heat. These generalizations are natural future work. Finally, $\tilde{\mathcal{D}}$ and $\bar\lambda_{\mathrm{rad}}$ are noisy to estimate, whereas $\tilde{\mathcal{V}}$ is stable and predicts $R$ well in our experiments.

\appendix

\section{Impact Statement}

This work develops a theoretical framework for understanding embedding norm dynamics in contrastive learning and derives a practical norm-aware similarity score. The primary applications are retrieval, calibration, and uncertainty analysis using existing pretrained models; no new models or datasets are released. We anticipate no direct negative societal impacts. Improved retrieval quality could modestly benefit downstream systems that rely on semantic search, with the usual caveats that such systems inherit biases present in pretraining data. The theory itself is foundational: it may inform future training procedures, regularization choices, and evaluation protocols for embedding models, potentially improving their reliability and interpretability.

\section{Additional Model Verifications}
\label{app:more_models_verification}

\begin{figure}[h]
  \centering
  \includegraphics[width=\linewidth]{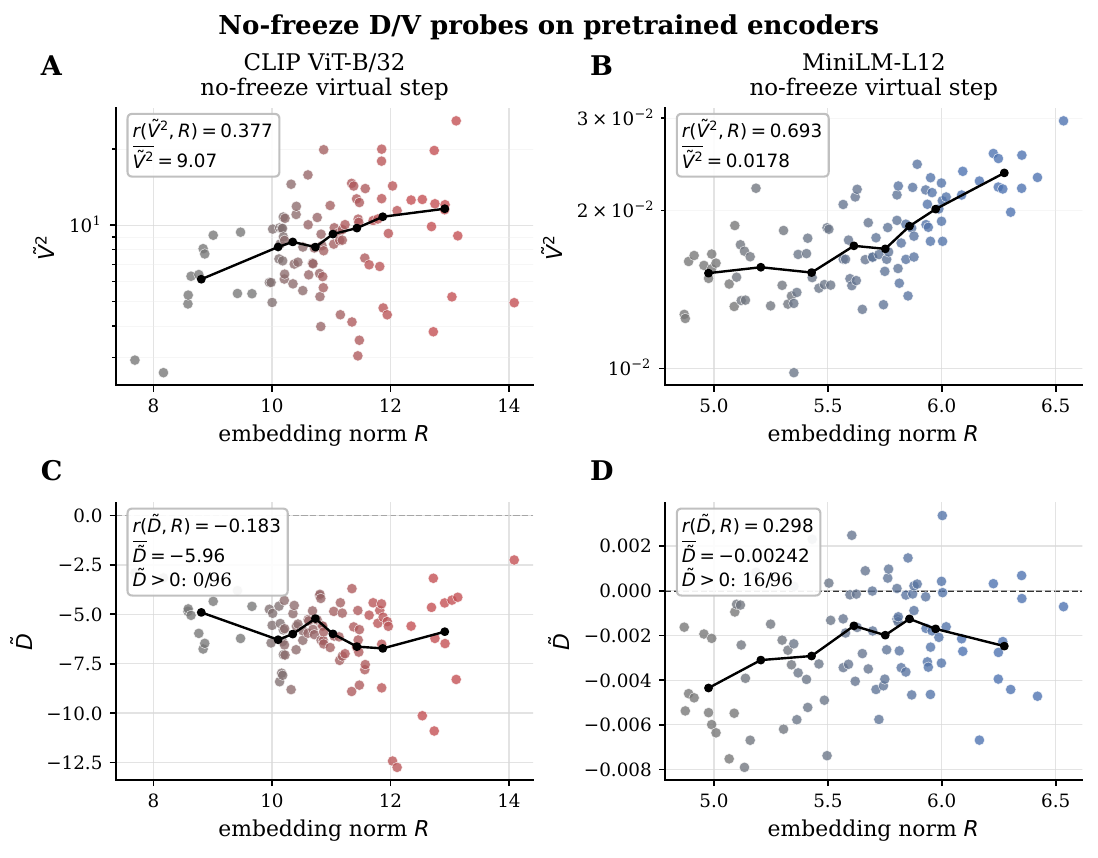}
  \caption{\textbf{$\tilde{\mathcal{V}}^2$ and $\tilde{\mathcal{D}}$ measurements.} Top row: per-token $\tilde{\mathcal{V}}^2$ and $\tilde{\mathcal{D}}$ values sorted by observed embedding norm for CLIP ViT-B/32 (left) and MiniLM-L12 (right). Bottom row: log-log relationship between $\tilde{\mathcal{V}}^2$ and norm, and $\tilde{\mathcal{D}}$ vs.\ $\tilde{\mathcal{V}}^2$ scatter.}
  \label{fig:additional_model_dv}
\end{figure}

We repeat the $\tilde{\mathcal{V}}$ and $\tilde{\mathcal{D}}$ measurement protocol from Section~\ref{sec:exp_dv_equation} without fine-tuning on two additional architectures---OpenAI CLIP ViT-B/32 (vision-language, 151M) and sentence-transformers/all-MiniLM-L12-v2 (text-only, 33M). For each model we select $n=96$ probe tokens spanning the full range of observed embedding norms, draw $M=64$ random minibatches of size $64$ from the MultiNLI entailment-pair probe objective described in Appendix~\ref{app:dv_measurement}, and compute the Monte Carlo estimators.
Figure~\ref{fig:additional_model_dv} displays the per-token values, and
Table~\ref{tab:additional_model_dv_summary} summarizes the statistics.

\begin{table}[h]
\centering
\caption{$\tilde{\mathcal{D}}$ and $\tilde{\mathcal{V}}^2$ summary statistics ($n=96$ probes each). Means reported $\pm$ SD. $\tilde{\mathcal{D}}<0$ indicates systematic inward radial drift. $^{***}p<0.001$, $^{**}p<0.01$, $^{*}p<0.05$ (two-sided $t$-test for Pearson $r$, $\mathrm{d.f.}=94$).}
\label{tab:additional_model_dv_summary}
\resizebox{\linewidth}{!}{%
\begin{tabular}{lccccc}
\toprule
Model & $\langle\tilde{\mathcal{V}}^2\rangle$ & $\langle\tilde{\mathcal{D}}\rangle$ & $r(\tilde{\mathcal{V}}^2, R)$ & $r(\tilde{\mathcal{D}}, R)$ & $r(\tilde{\mathcal{D}}, \tilde{\mathcal{V}}^2)$ \\
\midrule
CLIP ViT-B/32 & $9.07\pm4.09$ & $-5.96\pm1.79$ & $+0.377^{***}$ & $-0.183$ & $-0.580^{***}$ \\
MiniLM-L12 & $0.0178\pm0.0037$ & $-0.0024\pm0.0024$ & $+0.693^{***}$ & $+0.298^{**}$ & $+0.158$ \\
\bottomrule
\end{tabular}
}
\end{table}

As in the fine-tuned case, $\tilde{\mathcal{V}}^2$ predicts the norm (CLIP $r=0.377$, MiniLM $r=0.693$), though more weakly. This shows the effect is robust even without fine-tuning.

\section{Depth-Stratified TREC CAR Retrieval}
\label{app:trec_car}

To test whether norm discounting preferentially benefits queries at deeper specificity levels, we evaluate on TREC CAR~\cite{dietz2017trec} (Complex Answer Retrieval), where reported test-query depth buckets include page (depth~1, $22$ queries), section (depth~2, $1{,}012$ queries), subsection (depth~3, $685$ queries), and subsubsection (depth~4, $134$ queries). Deeper headings correspond to more specific subtopics. We sweep the document-norm exponent $\alpha$ in $\mathrm{score}(q,d)=\cos(q,d)\cdot\|d\|^\alpha$ and measure NDCG@10 gain over cosine ($\alpha=0$) at each depth. We compare CLIP~\cite{radford2021clip} variants and OpenCLIP~\cite{ilharco2021openclip} ViT-B/32 against text-only models (E5~\cite{wang2022e5}, BGE-large-v1.5~\cite{xiao2023cpack}, SBERT-MPNet~\cite{reimers2019sbert}, MiniLM-L6 and -L12~\cite{wang2020minilm}, DistilBERT~\cite{sanh2019distilbert}, ALBERT-small~\cite{lan2020albert}).

\begin{table}[t]
  \caption{NDCG@10 gain (\%) from norm-weighted retrieval on TREC CAR, by Wikipedia heading depth. Positive gain means the best norm-weighted score improves over cosine. Overall column shows gain and the optimal $\alpha$ across the evaluated queries.}
  \label{tab:car_depth}
  \centering
  \resizebox{\linewidth}{!}{%
  \begin{tabular}{llccccc}
    \toprule
    Model & Type & Overall (best $\alpha$) & d=1 (22q) & d=2 (1012q) & d=3 (685q) & d=4 (134q) \\
    \midrule
    \multicolumn{7}{c}{\textbf{Vision-language with projection head}} \\
    CLIP ViT-B/32 & VL+proj & \textbf{+32.3\%} ($-$0.75) & +19.6\% & +35.2\% & +27.1\% & +37.9\% \\
    CLIP ViT-B/16 & VL+proj & \textbf{+37.7\%} ($-$0.75) & +35.6\% & +39.1\% & +33.4\% & +43.3\% \\
    CLIP ViT-L/14 & VL+proj & \textbf{+44.3\%} ($-$1.25) & +18.7\% & +41.1\% & +48.2\% & +62.2\% \\
    OpenCLIP ViT-B/32 & VL+proj & +9.6\% ($-$1.00) & +23.5\% & +7.1\% & +14.3\% & +16.6\% \\
    \midrule
    \multicolumn{7}{c}{\textbf{VL without projection head (ablation)}} \\
    CLIP mean-pool & VL$-$proj & +7.9\%\textsuperscript{*} (+0.50) & +5.2\% & +8.4\% & +6.6\% & +16.8\% \\
    \midrule
    \multicolumn{7}{c}{\textbf{Text-only models}} \\
    E5-base-v2 & Text & +0.5\% ($-$0.10) & +3.3\% & 0.0\% & +1.1\% & +1.6\% \\
    BGE-large-v1.5 & Text & 0.0\% (+0.00) & +1.9\% & 0.0\% & +0.2\% & +1.1\% \\
    SBERT-MPNet & Text & +1.3\% ($-$0.30) & +12.4\% & +1.6\% & +1.3\% & +1.5\% \\
    MiniLM-L6 & Text & +1.4\% ($-$0.10) & +4.4\% & +1.4\% & +2.3\% & +0.3\% \\
    MiniLM-L12 & Text & +0.7\% ($-$0.20) & +3.2\% & +1.0\% & +0.9\% & +1.5\% \\
    DistilBERT & Text & +1.5\% ($-$0.10) & +3.3\% & +1.5\% & +0.8\% & +7.9\% \\
    ALBERT-small & Text & 0.0\% (+0.00) & +5.1\% & +0.3\% & 0.0\% & +2.0\% \\
    \bottomrule
  \end{tabular}
  }
  \vspace{0.2cm}
  {\footnotesize\textsuperscript{*}Gain uses \emph{positive} $\alpha$, i.e.\ boosting high-norm documents --- the opposite direction from VL+proj models, suggesting that the projection head is important for this specificity-norm signal.\par}
\end{table}

\begin{figure}[t]
  \centering
  \includegraphics[width=\linewidth]{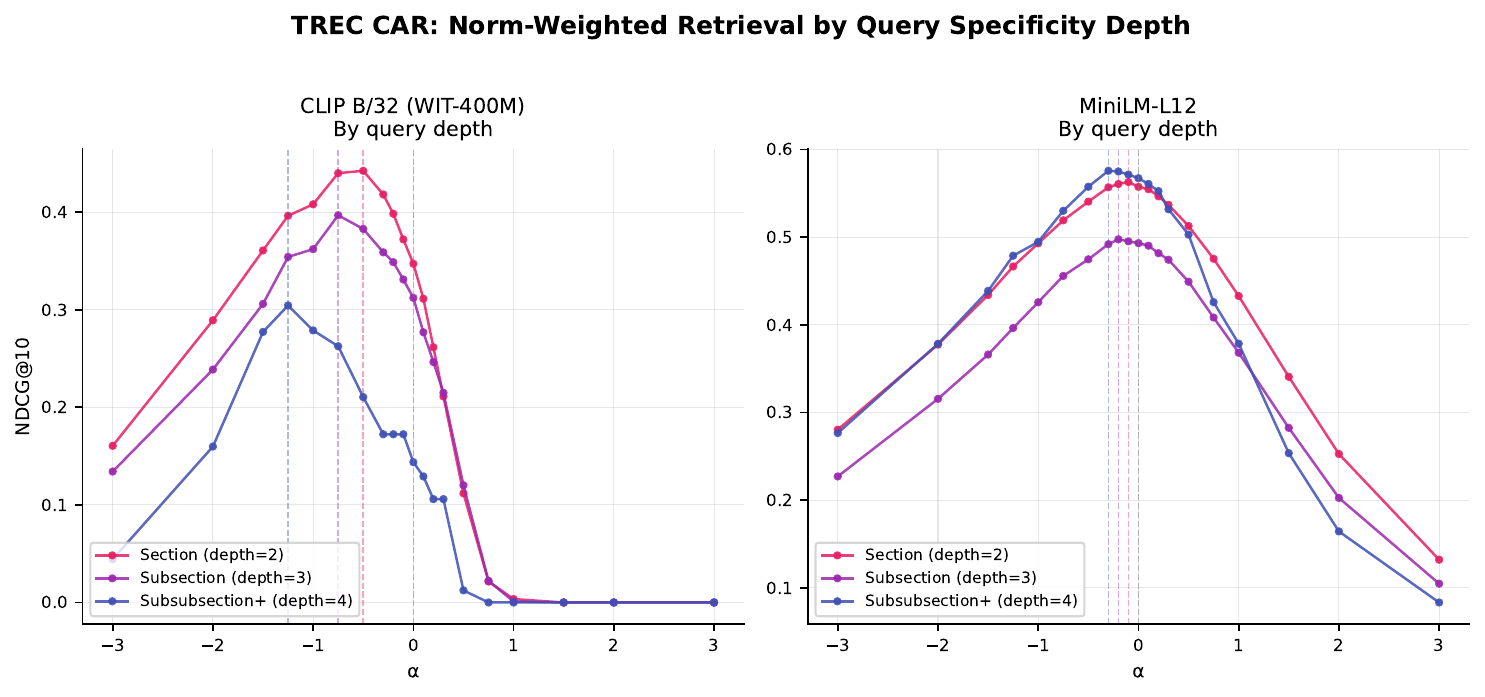}
  \caption{TREC-CAR depth-stratified norm-weighted retrieval. Each color represents a query specificity level (depth~2 = section, depth~3 = subsection, depth~4+ = subsubsection). CLIP benefits most from negative $\alpha$ at deeper heading levels; MiniLM-L12 shows no depth gradient.}
  \label{fig:car_depth}
\end{figure}

Table~\ref{tab:car_depth} and Figure~\ref{fig:car_depth} show a benchmark-specific split. All four vision-language models with projection heads gain substantially (+9.6\% to +44.3\%), and three of four select increasingly negative $\alpha$ as query depth increases from section (d=2) to subsubsection (d=4). Removing the projection head (CLIP mean-pool) flips the best exponent positive and removes the depth gradient. All seven text models stay near cosine ($\leq 1.5\%$ gain), with no systematic $\alpha$ pattern. These retrieval results stress-test, rather than repeat, the model-regime distinction measured in Section~\ref{sec:exp_dv_equation}.

\FloatBarrier

\section{Norm as a proxy for inverse specificity and model uncertainty}
\label{app:exp_specificity}

The theory identifies $\tilde{\mathcal{V}}$ with inverse specificity: broad, ambiguous, or polysemous inputs have larger transmitted update variance, while narrow and specific inputs have smaller $\tilde{\mathcal{V}}$. Section~\ref{sec:theory} suggests that the norm is also a proxy for inverse specificity when $\tilde{\mathcal{V}}$ increases with $R$. In the following experiments, we empirically investigate the relationship of both $\tilde{\mathcal{V}}$ and $R$ to semantic specificity through this inverse direction.

\paragraph{CIFAR-10H \texorpdfstring{$\tilde{\mathcal{V}}$}{V-tilde} multi-model experiment.}
We test whether $\tilde{\mathcal{V}}$ tracks inverse specificity and uncertainty on CIFAR-10H~\cite{peterson2019cifar10h}, which provides human soft-label distributions for 10000 CIFAR-10 images. We measure $\tilde{\mathcal{V}}$ for 256 norm-stratified probe images via 48 virtual contrastive SGD steps, across 14 vision-language models spanning OpenAI CLIP~\cite{radford2021clip}, OpenCLIP~\cite{ilharco2021openclip}, MetaCLIP~\cite{xu2024metaclip}, and SigLIP~\cite{zhai2023siglip}. For each probe we compute $H_{\text{model}}$ (zero-shot softmax entropy) and $H_{\text{human}}$ (entropy of the human soft-label distribution). Table~\ref{tab:vtilde_cifar10h} reports Pearson correlations. Across the 11 standard contrastive models, $r(\tilde{\mathcal{V}}^2, H_{\text{model}})$ is positive in every case and statistically significant in most. Correlations with human entropy are also generally positive, but weaker and less consistent. The norm correlation is often large, though not uniformly the largest across models. The three SigLIP variants (sigmoid rather than contrastive loss) show weaker and mixed correlations, consistent with the signal depending on the contrastive-training regime.

\begin{table}[t]
\caption{\textbf{$\tilde{\mathcal{V}}^2$ correlations across 14 vision-language models on CIFAR-10H.} $\tilde{\mathcal{V}}^2$ is positively correlated with model-internal uncertainty ($r(\tilde{\mathcal{V}}^2, H_{\text{model}})$) for all standard contrastive models and generally, but less consistently, correlated with human annotator uncertainty ($r(\tilde{\mathcal{V}}^2, H_{\text{human}})$). SigLIP models (marked $\dagger$) use sigmoid rather than contrastive loss and show weaker, mixed correlations.}
\label{tab:vtilde_cifar10h}
\centering
\small
\begin{tabular}{@{}l r r r @{}}
\toprule
Model & $r(\tilde{\mathcal{V}}^2, H_{\mathrm{model}})$ & $r(\tilde{\mathcal{V}}^2, H_{\mathrm{human}})$ & $r(\tilde{\mathcal{V}}^2, \|z\|)$ \\
\midrule
CLIP ViT-B/32            & $+0.234$\rlap{$^{***}$} & $+0.050$ & $+0.350$\rlap{$^{***}$} \\
CLIP ViT-B/16            & $+0.205$\rlap{$^{***}$} & $+0.147$\rlap{$^{*}$}   & $+0.179$\rlap{$^{**}$} \\
CLIP ViT-L/14@336px      & $+0.075$ & $+0.098$ & $+0.090$ \\
CLIP ViT-L/14            & $+0.052$ & $+0.010$ & $+0.364$\rlap{$^{***}$} \\
OpenCLIP ViT-B/32        & $+0.319$\rlap{$^{***}$} & $+0.210$\rlap{$^{***}$} & $+0.436$\rlap{$^{***}$} \\
OpenCLIP ViT-H/14        & $+0.177$\rlap{$^{**}$}  & $+0.178$\rlap{$^{**}$}  & $+0.126$\rlap{$^{*}$} \\
OpenCLIP ViT-L/14        & $+0.123$\rlap{$^{*}$}   & $+0.236$\rlap{$^{***}$} & $+0.259$\rlap{$^{***}$} \\
MetaCLIP ViT-B/16        & $+0.307$\rlap{$^{***}$} & $+0.087$ & $+0.495$\rlap{$^{***}$} \\
MetaCLIP ViT-H/14        & $+0.204$\rlap{$^{**}$}  & $+0.070$ & $+0.356$\rlap{$^{***}$} \\
MetaCLIP ViT-B/32        & $+0.190$\rlap{$^{**}$}  & $+0.161$\rlap{$^{**}$}  & $+0.397$\rlap{$^{***}$} \\
MetaCLIP ViT-L/14        & $+0.114$ & $+0.192$\rlap{$^{**}$}  & $+0.208$\rlap{$^{***}$} \\
SigLIP Base Patch16/384$\dagger$   & $+0.030$ & $+0.043$ & $+0.078$ \\
SigLIP Large Patch16/384$\dagger$  & $-0.029$ & $+0.099$ & $-0.172$\rlap{$^{**}$} \\
SigLIP SO400M Patch14/384$\dagger$ & $-0.066$ & $-0.026$ & $+0.192$\rlap{$^{**}$} \\
\bottomrule
\end{tabular}
\vspace{2pt}

\small $\dagger$ SigLIP uses sigmoid loss instead of contrastive. $^{***}p<0.001$, $^{**}p<0.01$, $^{*}p<0.05$ (two-sided $t$-test, $n=256$, $\mathrm{d.f.}=254$).
\end{table}

\FloatBarrier

\paragraph{Norm calibration via general/specific Q/A pairs.}
To further probe whether norm tracks inverse specificity at the answer level (rather than the concept level), we construct a synthetic Q/A dataset using a two-stage LLM pipeline. For each item, a generator LLM (DeepSeek-V4-Flash~\cite{deepseek2026}) produces a question, a general answer (broad background, definitions), and a specific answer (concrete mechanistic or quantitative details, same relevance). Another judge LLM validates each triple, requiring both answers to be equally relevant and correct, the specific to entail the general but not vice versa, and the specific to add concrete information. The final dataset contains 1060 validated triples spanning 17 domains.

We extract unnormalized embeddings for each answer across 10 models: mean-pool encoders MiniLM-L3~\cite{wang2020minilm}, E5-Large-v2~\cite{wang2022e5}, SBERT (all-mpnet)~\cite{reimers2019sbert}, BGE-M3~\cite{chen2024bge-m3}; vision-language CLIP ViT-B/32~\cite{radford2021clip}; and decoder models Qwen3-0.6B~\cite{qwen3_2025}, GTE-Qwen2-1.5B-instruct~\cite{li2023towards}, Jina-v3~\cite{sturua2024jina}, SFR-Mistral~\cite{meng2024sfr}, E5-Mistral-7B~\cite{wang2024e5mistral}. To control for the differing text lengths of the two answers, we apply the \emph{repeated-general protocol}: the general answer text is concatenated with itself before encoding so that the two answers have comparable token counts. For mean-pooled models this reduces the length confound, but duplicating text can still change contextual token states; for decoder models the 2$\times$ repetition alters the terminal context, so the comparison is less direct. The theory predicts $\|\text{specific}\| < \|\text{general}\|$ because the specific answer's narrower semantic content induces lower $\tilde{\mathcal{V}}$.

Table~\ref{tab:general_specific_norms} reports the results. For several encoder models, the specific answer has the smaller norm in a majority of pairs, consistent with the theory's prediction that narrower semantic content induces lower stochastic heat. The cosine sign fractions show model-dependent directional biases and should be read as a coarse relevance control rather than proof of equal relevance. Decoder model results are mixed, likely reflecting the repeated-general protocol's effect on last-token pooling. Overall, for some models the norm provides a signal that distinguishes specific from general answers without supervision.

\begin{table}[t]
\caption{Embedding norm and cosine similarity: specific vs.\ general answers
  under the repeated-general protocol ($n=1060$ pairs).
  $z = \sqrt{n}\,\bar\Delta / \sigma_\Delta$ (paired $z$-statistic);
  \%Norm$\downarrow$ = pairs where $\|\text{specific}\| < \|\text{general}_{\times 2}\|$;
  \%Cos$\downarrow$ = pairs where $\cos(q, \text{specific}) < \cos(q, \text{general})$;
  this reports the sign of the relevance-control difference, not its magnitude.
  Negative $z$ means specific answers have reliably smaller norms as predicted.}
\label{tab:general_specific_norms}
\centering
\small
\begin{tabular}{lccc}
\toprule
Model & $z$ & \%Norm$\downarrow$ & \%Cos$\downarrow$ \\
\midrule
\multicolumn{4}{l}{\textbf{Encoder (mean-pool)}} \\
\quad MiniLM-L3          & $-45.0$ & 93.5\% & 66.9\% \\
\quad E5-Large-v2        & $-52.7$ & 96.1\% & 85.4\% \\
\quad SBERT (all-mpnet)  & $-20.1$ & 73.2\% & 64.6\% \\
\quad BGE-M3             & $-12.5$ & 65.1\% & 74.0\% \\
\midrule
\multicolumn{4}{l}{\textbf{Vision-language (projection head)}} \\
\quad CLIP ViT-B/32      & $-6.3$  & 60.6\% & 87.1\% \\
\midrule
\multicolumn{4}{l}{\textbf{Decoder (last-token)}} \\
\quad Qwen3-0.6B         & $-14.4$ & 67.6\% & 59.6\% \\
\quad GTE-Qwen2-1.5B-instruct & $+0.0$  & 51.3\% & 48.7\% \\
\quad Jina-v3            & $+0.2$  & 51.6\% & 70.8\% \\
\quad SFR-Mistral        & $+36.4$ & 11.9\% & 71.3\% \\
\quad E5-Mistral-7B      & $+31.0$ & 18.0\% & 63.9\% \\
\bottomrule
\end{tabular}
\end{table}

\FloatBarrier

%% -----------------------------------------------------------------------
\section{Derivation of the Stationary Norm}
\label{app:stat_norm_derivation}

\paragraph{Main equation.}
Fix an input $x$, abbreviate $z_t=z_t(x)$, $z_t^+=z_t(x)+\Delta z_{\mathrm{task}}(x)$, $R=\|z_t\|$, and $\hat z_t=z_t/R$. Let $\Delta z_{\mathrm{task}}(x)$ denote the task update before weight decay, and write its conditional mean and fluctuation as
\begin{equation}
\bar\Delta_t(x):=\mathbb{E}[\Delta z_{\mathrm{task}}(x)\mid\theta_t],
\qquad
\varepsilon_t(x):=\Delta z_{\mathrm{task}}(x)-\bar\Delta_t(x),
\qquad
\mathbb{E}[\varepsilon_t(x)\mid\theta_t]=0.
\end{equation}
The squared-norm expansion is
\begin{equation}
\label{eq:norm_expansion}
\mathbb{E}_{B_t}[\|z_t+\Delta z_{\mathrm{task}}\|^2-\|z_t\|^2\mid\theta_t]
=2z_t^\top\bar\Delta_t+\|\bar\Delta_t\|^2
+\mathrm{Tr}\!\left(\mathrm{Cov}_{B_t}(\Delta z_{\mathrm{task}}\mid\theta_t)\right).
\end{equation}
Define the local effective radial damping of the post-task decay step by the conditional minibatch average
\begin{equation}
\label{eq:lambda_rad_def}
\mathbb{E}_{B_t}\!\left[
\|z_t^++\Delta z_{\mathrm{wd}}(x)\|^2-\|z_t^+\|^2
\,\middle|\,\theta_t
\right]
=:
-2\eta\lambda_{\mathrm{rad}}(x,t)R^2.
\end{equation}
Its stationary effective average is $\bar\lambda_{\mathrm{rad}}(x;R)$, as in Eq.~\eqref{eq:lambda_rad_stationary}. Equivalently, the damping definition includes the conditional average of
\begin{equation}
2\bigl(z_t+\Delta z_{\mathrm{task}}(x)\bigr)^\top\Delta z_{\mathrm{wd}}(x)
+\|\Delta z_{\mathrm{wd}}(x)\|^2
\end{equation}
inside the decay contribution. Define the rescaled task update $\Delta\tilde z_{\mathrm{task}}(x):=R\cdot\Delta z_{\mathrm{task}}(x)$ and the local state-conditioned moments
\begin{align}
\mu_t(x)&:=\mathbb{E}_{B_t}[\Delta\tilde z_{\mathrm{task}}(x)\mid\theta_t],\\
\tilde{\mathcal{D}}_t(x)&:=\hat z_t(x)^\top\mu_t(x),\\
\tilde{\mathcal{V}}_t(x)^2&:=\mathrm{Tr}\!\left(\mathrm{Cov}_{B_t}(\Delta\tilde z_{\mathrm{task}}(x)\mid\theta_t)\right),
\end{align}
and $\Sigma_t(x)^2:=\|\mu_t(x)\|^2+\tilde{\mathcal{V}}_t(x)^2$.
By definition, $\bar\Delta_t=\mu_t(x)/R$, so $z_t^\top\bar\Delta_t=\tilde{\mathcal{D}}_t(x)$ and $\|\bar\Delta_t\|^2+\mathrm{Tr}(\mathrm{Cov}_{B_t}(\Delta z_{\mathrm{task}}\mid\theta_t))=\Sigma_t(x)^2/R^2$ exactly.
Combining the task expansion with the post-task decay contraction gives
\begin{equation}
\label{eq:balance_condition}
\mathbb{E}_{B_t}\bigl[\|z_{t+1}\|^2-\|z_t\|^2\mid\theta_t\bigr]
=
-2\eta\lambda_{\mathrm{rad}}(x,t)R^2
+2\tilde{\mathcal{D}}_t(x)
+\frac{\Sigma_t(x)^2}{R^2}.
\end{equation}
The task--decay interaction is not discarded; it is included in $\lambda_{\mathrm{rad}}$ because the decay contribution is defined from the actual decay step after the task step. Averaging under the stationary condition $\|z_t(x)\|=R$, define
\begin{align}
\tilde{\mathcal{D}}_R(x)&:=\mathbb{E}_{\pi}\!\left[\tilde{\mathcal{D}}_t(x)\,\middle|\,\|z_t(x)\|=R\right],\\
\tilde{\mathcal{V}}_R(x)^2&:=\mathbb{E}_{\pi}\!\left[\tilde{\mathcal{V}}_t(x)^2\,\middle|\,\|z_t(x)\|=R\right],\\
\mathcal{M}_R(x)^2&:=\mathbb{E}_{\pi}\!\left[\|\mu_t(x)\|^2\,\middle|\,\|z_t(x)\|=R\right],\\
\Sigma_R(x)^2&:=\mathcal{M}_R(x)^2+\tilde{\mathcal{V}}_R(x)^2.
\end{align}
The radius-conditioned drift at radius $R$ is
\begin{equation}
\mathbb{E}_{\pi,B_t}\!\left[\|z_{t+1}\|^2-\|z_t\|^2\,\middle|\,\|z_t(x)\|=R\right]
=
-2\eta\bar\lambda_{\mathrm{rad}}(x;R)R^2+2\tilde{\mathcal{D}}_R(x)+\frac{\Sigma_R(x)^2}{R^2}.
\end{equation}
Let $R_t:=\|z_t(x)\|$. Stationarity of $\|z_t\|^2$ gives the outer-averaged identity
\begin{equation}
0
=
\mathbb{E}_{\pi}\!\left[
-2\eta\bar\lambda_{\mathrm{rad}}(x;R_t)R_t^2
+2\tilde{\mathcal{D}}_{R_t}(x)
+\frac{\Sigma_{R_t}(x)^2}{R_t^2}
\right].
\end{equation}
The closed-form norm follows after a narrow-distribution or zero-drift closure: choose $R_{\mathrm{eq}}$ as a radius where the conditional drift vanishes and assume the radius-conditioned quantities vary slowly near that radius. At $R=R_{\mathrm{eq}}$, using
\begin{equation}
\mathcal{D}_\star(x):=\frac{\tilde{\mathcal{D}}(x)}{\eta\bar\lambda_{\mathrm{rad}}(x)},
\qquad
\mathcal{V}_\star(x)^2:=\frac{\Sigma(x)^2}{\eta\bar\lambda_{\mathrm{rad}}(x)},
\end{equation}
the zero-drift equation divided by $\eta\bar\lambda_{\mathrm{rad}}(x)$ is
\begin{equation}
-2R_{\mathrm{eq}}^2+2\mathcal{D}_\star(x)+\frac{\mathcal{V}_\star(x)^2}{R_{\mathrm{eq}}^2}=0.
\end{equation}
Multiplying by $R_{\mathrm{eq}}^2$ gives the bi-quadratic
\begin{equation}
2R_{\mathrm{eq}}^4-2\mathcal{D}_\star(x)R_{\mathrm{eq}}^2-\mathcal{V}_\star(x)^2=0,
\end{equation}
whose nonnegative root is the zero-drift form in Theorem~\ref{thm:master_norm}.

\paragraph{NTK approximation.}
At step $t$, the task update satisfies $\Delta\theta\approx-\eta\nabla_\theta\mathcal{L}$, where $\mathcal{L}$ is the full averaged minibatch loss. The chain rule sends this update to the embedding of the fixed input $x$:
\begin{equation}
\Delta z_{\mathrm{task}}(x)
\approx
-\eta\sum_{k\in B_t}J_z(x)J_z(X_k)^\top\nabla_{z_t(X_k)}\mathcal{L}.
\end{equation}
The matrix $\mathcal K(x,X_k)=J_z(x)J_z(X_k)^\top$ is the Neural Tangent Kernel~\cite{jacot2018neural}. Since the loss depends on normalized embeddings, its source gradient is tangent and inverse-length scaled. Define
\begin{equation}
\label{eq:inv_scaling}
s_k:=\|z_t(X_k)\|\nabla_{z_t(X_k)}\mathcal{L},
\qquad
\nabla_{z_t(X_k)}\mathcal{L}=\frac{s_k}{\|z_t(X_k)\|},
\qquad
\hat z_t(X_k)^\top s_k=0.
\end{equation}
If the relevant NTK neighborhood has norms comparable to $R=\|z_t(x)\|$, then
\begin{equation}
\Delta z_{\mathrm{task}}(x)
\approx
\frac{1}{R}\Delta\tilde z_{\mathrm{task}}(x),
\qquad
\Delta\tilde z_{\mathrm{task}}(x)
:=
-\eta\sum_{k\in B_t}\mathcal K(x,X_k)s_k.
\end{equation}
Thus the abstract unit-norm kick in the main equation is the NTK-transmitted sum of tangent source gradients. The minibatch averaging is already contained in $s_k$ through the full-loss gradient.

\paragraph{Interpretable NTK moments.}
Let $\bar s(X):=\mathbb{E}[s_k\mid X_k=X]$ and $\Sigma_s(X):=\mathrm{Cov}(s_k\mid X_k=X)$. For exchangeable batch slots, neglecting cross-slot covariance terms induced by contrastive denominators,
\begin{align}
\mu_t(x)
&\approx
-\eta B\,\mathbb{E}_{X}\!\left[\mathcal K(x,X)\bar s(X)\right],\\
\tilde{\mathcal V}_t(x)^2
&\approx
\eta^2 B\,\mathrm{Tr}\!\left(
\mathbb{E}_{X}\!\left[\mathcal K(x,X)\Sigma_s(X)\mathcal K(x,X)^\top\right]
+\mathrm{Cov}_{X}\!\left(\mathcal K(x,X)\bar s(X)\right)
\right).
\end{align}
This is an interpretive weak-coupling approximation, not the estimator used in the experiments. The measurement protocol in Appendix~\ref{app:dv_measurement} estimates the task-update moments directly by virtual gradient steps, so it does not require the NTK covariance approximation.
The first line says the mean drift is the summed NTK-transmitted source update. The second line says stochastic heat grows when the NTK neighborhood of $x$ sends diverse tangent votes; in the weak-mean regime it reduces to the operator energy
\begin{equation}
\tilde{\mathcal V}_t(x)^2
\approx
\eta^2 B\,\mathbb{E}_{X}\!\left[
s(X)^\top\mathcal K(x,X)^\top\mathcal K(x,X)s(X)
\right].
\end{equation}

Radial drift depends on the sign of the transmitted mean vote:
\begin{equation}
\tilde{\mathcal D}_t(x)
=
\hat z_t(x)^\top\mu_t(x)
\approx
-\eta B\,\mathbb{E}_{X}\!\left[\hat z_t(x)^\top\mathcal K(x,X)\bar s(X)\right].
\end{equation}
Move the NTK onto the radial test direction. Since $\bar s(X)$ is tangent to the unit sphere at $X$, only the tangent projection matters:
\begin{equation}
a_K(x,X)
:=
P_{\hat{z}_t(X)}^\perp\mathcal{K}(x,X)^\top\hat{z}_t(x),
\qquad
P_{\hat{z}_t(X)}^\perp=I-\hat{z}_t(X)\hat{z}_t(X)^\top .
\end{equation}
Thus
\begin{equation}
\tilde{\mathcal D}_t(x)
\approx
\eta B\,\mathbb{E}_{X}\!\left[a_K(x,X)^\top(-\bar s(X))\right].
\end{equation}
Let $\varphi_K(x,X)$ be the angle between $\mathcal K(x,X)^\top\hat z_t(x)$ and $\hat z_t(X)$, and let $\psi_K(x,X)$ be the angle between $a_K(x,X)$ and $-\bar s(X)$. Since
\begin{equation}
\|a_K(x,X)\|
=
\|\mathcal K(x,X)^\top\hat z_t(x)\|\sin\varphi_K(x,X),
\end{equation}
the radial drift has the projected full-NTK form
\begin{equation}
\label{eq:drift_exact_projection}
\tilde{\mathcal D}_t(x)
\approx
\eta B\,\mathbb{E}_{X}\!\Big[
\underbrace{\|\mathcal K(x,X)^\top\hat z_t(x)\|}_{\text{Transmission}}
\underbrace{\sin\varphi_K(x,X)}_{\text{Concept direction alignment}}
\underbrace{\|\bar s(X)\|}_{\text{Update size}}
\underbrace{\cos\psi_K(x,X)}_{\text{Update alignment}}
\Big].
\end{equation}
Positive drift needs strong transmission, a large tangent component, a large source update, and positive alignment. Negative alignment compresses the norm; high variance without coherent alignment heats the embedding without creating outward drift.

%% -----------------------------------------------------------------------
\section{Variance Decomposition of the Transmitted Update}
\label{app:variance_decomp}

Let $\bar{s}(x):=\mathbb{E}[s_k\mid X_k=x]$ and $\Sigma_s(x):=\mathrm{Cov}(s_k\mid X_k=x)$. By the law of total covariance applied to the scale-normalized transmitted vector $\mathcal{K}(x_\ast,X_k)s_k$:
\begin{equation}
\mathrm{Cov}\!\left(\mathcal{K}(x_\ast,X_k)s_k\right)
= \underbrace{\mathbb{E}_{X}\!\left[\mathcal{K}(x_\ast,X)\Sigma_s(X)\mathcal{K}(x_\ast,X)^\top\right]}_{\text{intra-batch covariance}}
+ \underbrace{\mathrm{Cov}_{X}\!\left(\mathcal{K}(x_\ast,X)\bar{s}(X)\right)}_{\text{index-sampling covariance}}.
\end{equation}
Taking the trace separates residual batch noise (first term, noise after conditioning on which input is drawn) from index-sampling noise (second term, noise from drawing different inputs with different conditional mean gradients):
\begin{align}
\mathrm{Tr}\!\left(\mathrm{Cov}\!\left(\mathcal{K}(x_\ast,X_k)s_k\right)\right)
&=\mathbb{E}_{X}\!\left[\mathrm{Tr}\!\left(\mathcal{K}(x_\ast,X)\Sigma_s(X)\mathcal{K}(x_\ast,X)^\top\right)\right] \nonumber\\
&\quad+\mathrm{Tr}\!\left(\mathrm{Cov}_{X}\!\left(\mathcal{K}(x_\ast,X)\bar{s}(X)\right)\right).
\end{align}
With $Y_k:=\mathcal K(x_\ast,X_k)s_k$, the scale-normalized task update is $-\eta\sum_{k=1}^B Y_k$ under the full-loss-gradient convention used above. If the $Y_k$ were independent draws, the covariance of the sum would be $B$ times the single-slot covariance above. For InfoNCE-style losses, however, the $Y_k$ are coupled through the shared softmax denominators; the exact covariance also includes cross-slot terms,
\begin{equation}
\mathrm{Cov}\!\left(\sum_{k=1}^B Y_k\right)
=
B\,\mathrm{Cov}(Y_1)
+B(B-1)\,\mathrm{Cov}(Y_1,Y_2),
\end{equation}
where the second covariance denotes the covariance between two distinct exchangeable slots. The main-text approximation neglects this cross-slot term. The combined second-order task heat also includes the squared mean. With $Y(X):=\mathcal K(x_\ast,X)s(X)$,
\begin{equation}
\mathbb{E}\!\left[\left\|\sum_{k\in B_t}Y(X_k)\right\|^2\right]
\approx
B\,\mathbb{E}_{X}\!\left[\|Y(X)\|^2\right]
+
B(B-1)\left\|\mathbb{E}_{X}\!\left[Y(X)\right]\right\|^2.
\end{equation}
The first term captures stochastic heating from per-slot fluctuations; the second captures coherent drift and dominates for hub concepts with large $\|\mathbb{E}[\mathcal{K}s]\|$. Because $s_k$ is defined using the averaged minibatch loss, its magnitude already carries the usual minibatch-averaging scale. The approximation Eq.~\eqref{eq:energy_scaling} in the main text keeps the first term in the weak-mean regime.

%% -----------------------------------------------------------------------

%% -----------------------------------------------------------------------
\section{Optimizer Artifacts: Momentum}
\label{app:optimizer}

\paragraph{Momentum smoothing of stochastic heat.}
Momentum-style optimizers replace a raw stochastic update $U_t$ by an exponential moving average $m_t=\beta m_{t-1}+(1-\beta)U_t$. Modelling $U_t$ as i.i.d.\ with mean $\bar U$ and covariance $\Sigma_U$, the stationary moving average satisfies $\mathbb{E}[m_\infty]=\bar U$ and
\begin{equation}
\mathrm{Cov}(m_\infty)=\frac{1-\beta}{1+\beta}\Sigma_U,
\end{equation}
\textit{Derivation.} Unrolling the recursion gives $m_\infty = (1-\beta)\sum_{k=0}^\infty \beta^k U_{t-k}$, so by independence of the $U_t$,
\[
  \mathrm{Cov}(m_\infty) = (1-\beta)^2 \sum_{k=0}^\infty \beta^{2k}\,\Sigma_U
  = \frac{(1-\beta)^2}{1-\beta^2}\,\Sigma_U = \frac{1-\beta}{1+\beta}\,\Sigma_U.
\]
So at stationarity ($\bar U\approx 0$) momentum rescales stochastic heating by $(1-\beta)/(1+\beta)$ without changing which concepts receive larger transmitted variance.

\section{Experimental Details}
\label{app:exp_details}

\paragraph{Synthetic datasets.}
The synthetic experiments use artificial concept tokens paired with positive strings drawn from diverse template families representing semantically coherent sentence variations (e.g., ``researchers compared \textit{pig} and \textit{swine} in animal behavior studies''). The $\lambda$-scaling experiment uses 100 positives per concept; the frequency-independence experiment varies the number of positives over $\rho\in\{5,20,100,500,2000\}$ while holding semantic content fixed. These token types are treated as natural-vocabulary words (not special tokens), added to the tokenizer vocabulary and resized in the embedding table.

Training uses \texttt{MultipleNegativesRankingLoss} from \texttt{sentence-transformers}~\cite{reimers2019sbert}, SGD with weight decay (SGDW), batch size 256, and learning rate $10^{-3}$. Each $\lambda$ value is trained independently to norm convergence (early stop when $\max_x |\Delta\|z(x)\|| < 5 \times 10^{-2}$ and relative change $< 1.5 \times 10^{-2}$ for 5 consecutive checks every 5 epochs, or when the step cap of 700 updates is reached). All experiments use \texttt{distilbert-base-uncased}~\cite{sanh2019distilbert} as the base model.

\paragraph{$\tilde{\mathcal{D}}$ and $\tilde{\mathcal{V}}$ measurement.}
\label{app:dv_measurement}
We estimate $\tilde{\mathcal{D}}(x)$ and $\tilde{\mathcal{V}}(x)^2$ via Monte Carlo gradient probing. For $M$ random minibatches drawn from the relevant training or probe pairs, we compute the contrastive loss gradient, apply a virtual gradient step of size $\delta$, record the full pre-normalization embedding displacement vector $\Delta z^{(m)} = z'^{(m)} - z$, then restore the parameters. We report the update-scale unit-norm displacement
\begin{equation}
\Delta\tilde{z}^{(m)} := \Delta z^{(m)} \cdot \|z\|,
\end{equation}
which approximates one draw from the update-scale task displacement corresponding to $\Delta\tilde{z}_\mathrm{task}$. Dividing by $\delta$ gives the corresponding first-order task velocity; in correlation and scaling tests, the common step-size factor is handled consistently across probes. Let $\bar{\Delta\tilde{z}} = \frac{1}{M}\sum_{m=1}^{M} \Delta\tilde{z}^{(m)}$ be the sample mean vector. The estimators are
\begin{align}
\hat{\tilde{\mathcal{D}}}(x) &= \hat{z}(x)^\top \bar{\Delta\tilde{z}}, \\
\hat{\tilde{\mathcal{V}}}(x)^2 &= \frac{1}{M-1}\sum_{m=1}^{M} \|\Delta\tilde{z}^{(m)} - \bar{\Delta\tilde{z}}\|^2.
\end{align}
The first line projects the mean displacement onto the radial direction; the second is the trace of the sample covariance of the full $d$-dimensional displacement, matching the theoretical definition $\tilde{\mathcal{V}}^2 = \mathrm{Tr}(\mathrm{Cov}(\Delta\tilde{z}_\mathrm{task}))$ from Theorem~\ref{thm:master_norm}. Estimating $\tilde{\mathcal{D}}$ from the scalar norm change as $\|z\|\Delta R$ (or $\|z\|\Delta R/\delta$ under the velocity convention) recovers $\hat{z}^\top\bar{\Delta\tilde{z}}$ to first order and is equivalent; estimating $\tilde{\mathcal{V}}^2$ from the variance of the scalar $\|z\|\Delta R$ would instead give $\hat{z}^\top\mathrm{Cov}(\Delta\tilde{z})\hat{z}$, the radial component only, and is \emph{not} used here. Because the effective post-task radial damping varies across concept positions and training steps, we estimate it by measuring the squared-norm contribution of the decay step after the task step; details follow Eq.~\eqref{eq:lambda_rad_stationary}.

\paragraph{Pretrained-model virtual objectives.}
For pretrained models, these measurements are not intended to reconstruct the inaccessible original pretraining update distribution. They use standardized surrogate contrastive objectives to define local virtual-gradient probes. For the CLIP/MiniLM token probes in Appendix~\ref{app:more_models_verification}, the objective uses MultiNLI entailment pairs: premises are anchors, hypotheses are positives, and other positives in the minibatch act as negatives under a normalized in-batch contrastive loss. The virtual step updates all model parameters and is immediately undone, so no fine-tuning persists. For the CIFAR-10H experiment below, the objective is an image-to-class contrastive probe using the ten CIFAR class prompts ``a photo of a \{class\}'' as text prototypes.

\paragraph{CIFAR-10H uncertainty experiment.}
For the multi-model CIFAR-10H experiment in Appendix~\ref{app:exp_specificity}, human uncertainty per image is defined as the entropy of the soft label distribution over 10 classes. We compute model uncertainty from zero-shot class probabilities and estimate $\tilde{\mathcal{V}}$ using $M=48$ virtual minibatches of size $32$ under the CIFAR image-to-class objective above; the virtual steps are undone, so no fine-tuning or adaptation occurs.

\paragraph{Retrieval experiments.}
For retrieval experiments, embeddings are precomputed once for each model and reused across all norm exponents $\alpha$.

\paragraph{LLM-generated general--specific QA dataset.}
\label{app:general_specific_qa}
The controlled specificity-selection experiment uses a dataset of $1{,}060$ query--answer pairs generated by an LLM and curated with an LLM judge. Each item consists of a factual query, a \emph{general} answer, and a \emph{specific} answer. The general answer conveys the essential fact in plain terms; the specific answer adds concrete mechanism, numerical thresholds, named entities, or process detail absent from the general answer. Items are drawn from 17 domains: medicine ($150$), law ($143$), computer systems ($124$), biology ($122$), chemistry ($107$), physics ($104$), machine learning ($90$), economics ($54$), mathematics ($36$), astronomy ($30$), climate science ($27$), history ($26$), public policy ($15$), statistics ($13$), engineering ($11$), geography ($5$), and finance ($3$).

We generated the triples with an OpenAI-compatible chat-completions model using a fixed system prompt, temperature $0.65$, and \texttt{max\_tokens=8192}; a separate JSON-based judge pass accepted a candidate only when both answers were relevant and correct, carried the same relevance label, the specific answer entailed the general answer but not vice versa, the specific answer added concrete information, and the pair was non-contradictory. This acceptance criterion is designed to make the two answers differ primarily in specificity rather than in relevance or correctness.

To control for differing text lengths between the two answers, we adopt a \emph{repeated-general} protocol: the general answer is concatenated with itself before encoding, so that both answers have comparable token counts. For mean-pooled models this reduces the length confound, but duplicating text can still change contextual token states. For last-token-pooled (decoder) models, the repetition affects the last token, and we report results with this stronger caveat.

For each query--answer pair, we encode the query $q$, the repeated general answer $a_g$, and the specific answer $a_s$ with a frozen embedding model. We then compute the paired $z$-statistic $z = \sqrt{n} \cdot \overline{\Delta} / \sigma_\Delta$ where $\Delta_i = \|a_s^{(i)}\| - \|a_g^{(i)}\|$, as well as the fraction of pairs where the specific answer has the smaller norm. As a relevance control, we also compute the sign of $\cos(q, a_s) - \cos(q, a_g)$; this sign statistic detects directional bias but does not by itself establish small cosine-difference magnitudes.

The experiment is run across 10 embedding models spanning encoder-only (mean-pooled), vision-language, and decoder-only architectures. All embeddings are extracted without fine-tuning.

\printbibliography

\end{document}